\pdfoutput=1

\documentclass[11pt]{article}
\usepackage{ACL2024}

\usepackage{times}
\usepackage{latexsym}
\usepackage{algorithm}
\usepackage{algorithmic}
\usepackage{url}
\usepackage[T1]{fontenc}

\usepackage[utf8]{inputenc}
\usepackage{amssymb}

\usepackage{microtype}

\usepackage{inconsolata}
\usepackage{booktabs}
\usepackage{multirow}
\usepackage{float}
\usepackage{adjustbox}
\usepackage{caption}
\usepackage{subcaption}
\usepackage{comment}
\usepackage{siunitx}
\usepackage{amsmath}
\usepackage{mathtools}
\usepackage{tcolorbox}
\usepackage{caption}
\usepackage{varwidth}
\usepackage{ragged2e}
\usepackage{longtable}
\usepackage{tabularx}

\definecolor{Celeste}{RGB}{203, 239, 237}
\definecolor{Salmone}{RGB}{249, 210, 218}
\definecolor{Beige}{RGB}{246, 231, 205}
\definecolor{Lilletto}{RGB}{239, 227, 249}

\definecolor{Verdetto}{RGB}{170, 230, 170}
\definecolor{Rossetto}{RGB}{230, 170, 170}

\usepackage[inline]{enumitem}
\usepackage{appendix}
\newcommand{\resource}{\textsc{LLM-Oasis}}
\newcommand{\numInstances}{81k}
\newcommand{\numInstancesPrecise}{81,275}

\newcommand{\argmax}{\mathop{\mathrm{argmax}}}

\title{Truth or Mirage? \raisebox{-0.25\height}{\includegraphics[scale=0.075]{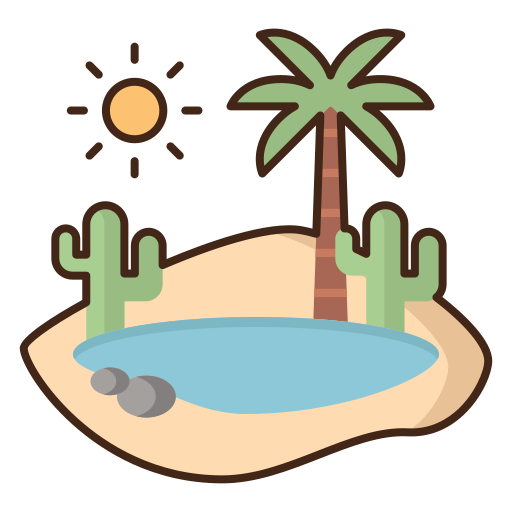}} \\Towards End-To-End Factuality Evaluation with \textsc{LLM-Oasis} \space}

\author{Alessandro Scir\`e\thanks{Equal contribution.}$^{1,2}$ \quad ~Andrei Stefan Bejgu$^{*1,2}$ \quad Simone Tedeschi$^{1,2}$\\{\bf Karim Ghonim$^{2}$}\quad {\bf Federico Martelli$^{2}$}\quad {\bf Roberto Navigli$^{2}$}\\\\
\qquad Babelscape, Italy  \qquad \qquad ~~~~~
          Sapienza University of Rome\\
   $^1$\texttt{lastname@babelscape.com} \qquad $^2$\texttt{\{first.lastname\}@uniroma1.it}}

\begin{document}
\maketitle
\begin{abstract}
After the introduction of Large Language Models (LLMs), there have been substantial improvements in the performance of Natural Language Generation (NLG) tasks, including Text Summarization and Machine Translation. However, LLMs still produce outputs containing hallucinations, that is, content not grounded in factual information.
Therefore, developing methods to assess the factuality of LLMs has become urgent. 
Indeed, resources for factuality evaluation have recently emerged. Although challenging, these resources face one or more of the following limitations: i) they are tailored to a specific task or domain; ii) they are limited in size, thereby preventing the training of new factuality evaluators, iii) they are designed for simpler verification tasks, such as claim verification.  
To address these issues, we introduce \resource, to the best of our knowledge the largest resource for training end-to-end factuality evaluators. \resource~is constructed by extracting claims from Wikipedia, falsifying a subset of these claims, and generating pairs of factual and unfactual texts. We then rely on human annotators to both validate the quality of our dataset and to create a gold standard test set for benchmarking factuality evaluation systems.
Our experiments demonstrate that \resource~presents a significant challenge for state-of-the-art LLMs, with GPT-4o achieving up to 60\% accuracy in our proposed end-to-end factuality evaluation task, highlighting its potential to drive future research in the field.
\end{abstract}

\section{Introduction}
In recent years, generative approaches in NLP have demonstrated remarkable results, achieving state-of-the-art performance across various tasks. This progress has been particularly notable with the advent of Large Language Models (LLMs), which have revolutionized the field, driving advancements in many tasks, including Text Summarization~\cite{news_summ_gpt, pu2023summarization, Zhang2023BenchmarkingLL}, Machine Translation~\cite{alves2024tower, zhang2023prompting, wang-etal-2023-document-level}, and Question Answering~\cite{kamalloo-etal-2023-evaluating, RASOOL2024100083}.
However, a critical challenge remains as LLMs' outputs still contain hallucinations, i.e. content that cannot be grounded in any pre-existing knowledge~\cite{tonmoy2024comprehensive, tam2022evaluating}. Compounding the problem, LLMs generate highly-fluent texts~\cite{wang2023elementaware}, which may mislead users into trusting their factual accuracy. Therefore, developing modeling strategies to mitigate this issue and creating tools to detect and correct hallucinations has become urgent.
\begin{figure*}[t!]
    \centering
\includegraphics[scale=0.7]{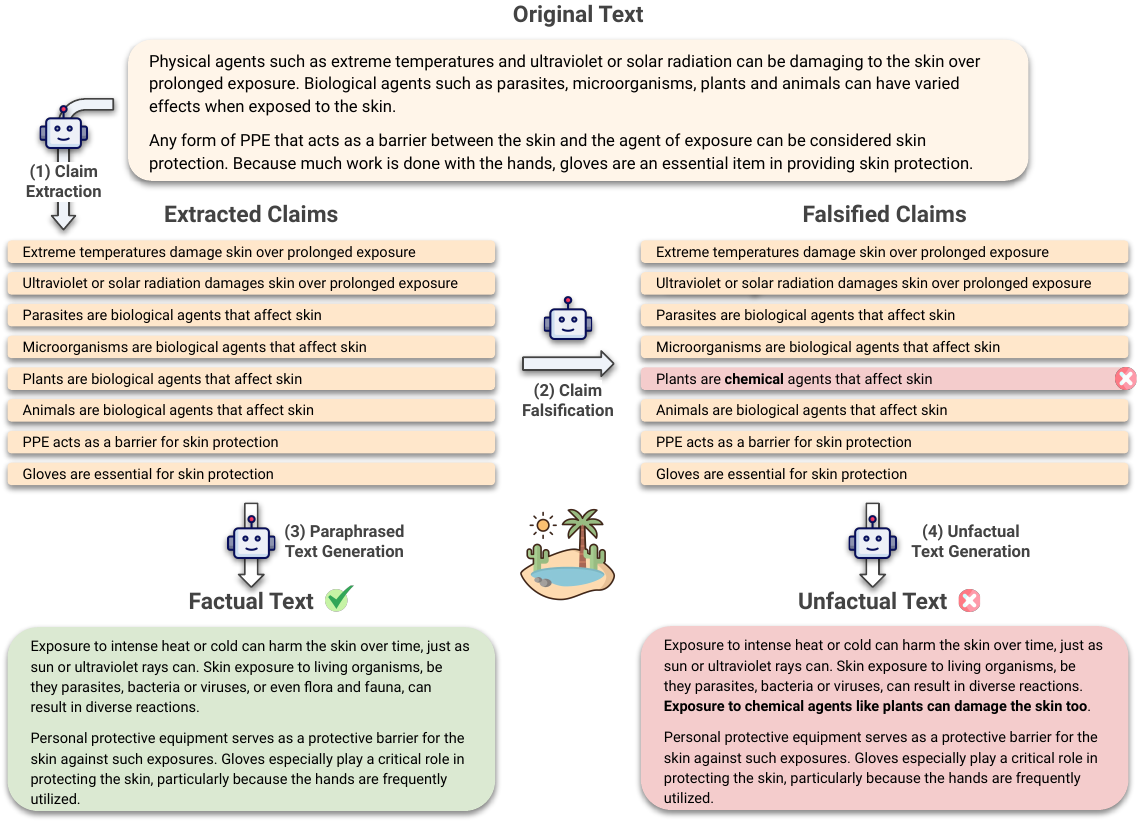}
    \caption{Pipeline for the creation of \resource. Given a passage from a Wikipedia page (original text on top), we task an LLM to: extract a list of atomic claims (1), falsify one of the extracted claims (2), and then, given the two sets of claims, produce a paraphrase of the original text (3), and an alternative version featuring the unfactual information (4).}
    \label{fig:falsepedia}
\end{figure*}
In this work, we focus on the problem of factuality evaluation, that is, the task of checking the factual accuracy of a machine-generated text. Previous research has proposed various resources to address this task. Although challenging, even for LLM-based factual reasoners, these resources are designed for specific settings, such as text summarization of news ~\cite{laban2021summac, tang-etal-2023-understanding}, books~\cite{scirè2024fenice}, and dialogues~\cite{tang2024tofueval}, among others. These benchmarks, while representative in their respective domains and tasks, often present peculiarities, which may lead to a lack of generalizability across different settings. A more general resource, pairing claims with evidence from Wikipedia is FEVER~\cite{thorne-etal-2018-fever}; however, its applicability is limited by its focus on claim verification, which involves assessing the veracity of individual facts. This formulation is not well-suited to real-world scenarios, where texts typically contain multiple facts, thereby preventing the development of end-to-end factuality evaluation systems.
These limitations highlight the need for a resource that is, not restricted to a specific domain or task, offering broader applicability and enabling the design of complete factuality evaluation approaches.
In this context, we introduce \resource, a large-scale resource for end-to-end factuality evaluation, created by extracting and falsifying information from Wikipedia pages. The overall process is depicted in Fig. \ref{fig:falsepedia}.
As a result, we obtain \numInstances~$\langle \text{factual}, \text{unfactual} \rangle$ pairs that are suitable to train end-to-end factuality evaluation systems.
Additionally, we setup a human annotation process to: i) create a gold standard for the factuality evaluation task, useful for benchmarking LLMs, and ii) validate the quality of the proposed data creation pipeline.
Additionally we issue two tasks, namely \textit{end-to-end factuality evaluation} and \textit{evidence-based claim verification} to benchmark LLMs. Our experiments demonstrate that our resource is challenging 
even for state-of-the-art models, both in zero-shot and Retrieval Augmented Generation \cite[RAG]{lewis2021retrievalaugmentedgenerationknowledgeintensivenlp} settings, with GPT-4o achieving an accuracy of 60\% and 68\%, respectively. 

In summary, our contributions are the following:
\begin{itemize}
    \vspace{-0.5mm}
    \item We introduce \resource, to the best of our knowledge the largest resource for end-to-end factuality evaluation, obtained by falsifying claims extracted from Wikipedia;
    \item Our resource enables two tasks to challenge current LLMs to detect factual inconsistencies in both short and long texts;
    \item We propose a gold standard benchmark, resulting from a human annotation process, to evaluate models on the proposed tasks;
    \item Our experiments demonstrate that our benchmark presents a significant challenge for LLMs, with smaller specialized models trained on \resource~achieving competitive performance.

\end{itemize}

Although we selected Wikipedia as the basis for our resource, we emphasize that our methodology can be potentially adapted to any other corpus in any domain or language, as the only requirement is access to a collection of raw texts. In the hope of fostering research in factuality evaluation, we release our resource and code at \url{https://github.com/Babelscape/LLM-Oasis}.

\section{Related Work}

Previous studies for factuality evaluation have focused on assessing \textit{factual consistency}, i.e., the extent to which a generated text is grounded in a source document. 
Resources for this task typically include human annotations that indicate whether a generated text accurately reflects the original document's facts. However, many of these works are tailored for specific tasks and domains, such as the assessment of factual consistency in summaries of news~\cite{fabbri2021summeval, tang-etal-2023-understanding, pagnoni2021understanding}, books~\cite{scirè2024fenice}, and dialogues~\cite{tang2024tofueval}.
Moreover, they are based on the assumption that the source of knowledge required for the verification is always available (e.g., the source document). This is not the case for the more general \textit{factuality evaluation} task, in which a text in natural language must be verified regardless of the availability of the evidence, potentially requiring information retrieval techniques.

The first contribution towards general-purpose factuality evaluation dates back to FEVER~\cite[Fact Extraction and VERification]{thorne-etal-2018-fever}, which pairs claims with evidence retrieved from Wikipedia. The FEVER dataset comprises 185,445 human-generated claims, created by modifying sentences extracted from Wikipedia and subsequently verified without knowledge of the original sentences. The claims are classified as \texttt{Supported}, \texttt{Refuted}, or \texttt{NotEnoughInfo}, and for the first two categories, annotators also recorded the sentence(s) forming the necessary evidence for their judgment. Although challenging, FEVER presents limitations due to its focus on fact verification, which involves checking the veracity of individual claims. This focus is hardly adaptable to real-world scenarios, where texts to verify usually feature multiple claims.
Additionally, FEVER's annotation effort is limited to a relatively-small subset of 10k English Wikipedia pages.

More recently, multiple studies introduced strategies to generate synthetic instances for factuality evaluation. Notably, \citet{muhlgay-etal-2024-generating} introduced FACTOR, a framework to generate factuality benchmarks by prompting an LLM to produce factual and unfactual completions given a prefix text. FACTOR includes 4,266 instances of $\langle$prefix, completion$\rangle$ pairs, each accompanied by a factuality label. Along the same lines, by introducing FELM, ~\citet{chen2023felm} provide 847 LLM outputs focused on different types of knowledge, such as World Knowledge, Math, and Reasoning with human-made factuality annotations. While valuable for benchmarking LLMs, the limited size of these resources prevents them from being used to train new factuality evaluators.

With \resource, we differentiate from previous studies by introducing a large-scale, task-agnostic resource covering a wide range of domains from Wikipedia.
Specifically, \resource~enables the task of end-to-end factuality evaluation, namely the more realistic scenario that involves the verification of raw text in natural language. Notably, texts falling under this setting, always go beyond individual sentences, inherently posing a more complex challenge to the systems.
Additionally, to the best of our knowledge, it is the largest resource for this task, featuring 162,550 passages in natural language and 681,201 claims, which can be verified against knowledge from \numInstancesPrecise~Wikipedia pages covering a broad set of domains. 
Finally, we reserve a manually-curated subset for this task, consisting of approximately 2k instances, and use it to benchmark several state-of-the-art models.







\begin{table*}[t!]
\begin{tabular}
{|p{0.95\textwidth}|}
\hline 
\vspace{1mm}
\textbf{Input:} Wikipedia Passage of $K$ sentences
\vspace{3mm}\\
\hline \\

\textbf{Instructions:} Execute the following steps: \\\\
\colorbox{Celeste}{\textbf{Step 1 - Claim extraction:}} From the input passage, extract a comprehensive set of claims. These claims must be atomic, i.e. semantically-coherent pieces of text that do not require further subdivision, and self-contained, i.e. not requiring additional context to be verified. Note that each claim must be short, using 15 words at most. Do not use "..." to truncate them. The ordering of the extracted claims must follow the logical flow expressed in the original text. Use a noun as the subject in the claim (avoid pronouns). All the claims that are featured in the input text must be reported in the list. \\\\
\colorbox{Salmone}{\textbf{Step 2 - Claim falsification:}} From the output of Step 1, subtly alter one claim, in order to introduce a critical factual inaccuracy. Such claim must be the most relevant for the input text. It is forbidden to change dates, years, numbers and person/location/organization/etc. names. It is also forbidden to provide naive negative transformations of verbs, e.g., was -> was not, did -> did not. This step, i.e., Step 2, returns a pair containing the altered claim along with the original one. \\\\
\colorbox{Beige}{\textbf{Step 3 - Factual text generation:}} From the output of Step 1, generate a text. Note that this text must be a paraphrase of the original provided text, i.e. a new text that should overlap as little as possible with the original, while preserving the meaning. The generated text must follow the same logical flow as the ordering of the extracted claims. \\\\
\colorbox{Lilletto}{\textbf{Step 4 - Unfactual text generation:}} Generate a text from the final set of claims (original unaltered + altered) i.e. the output of Step 3. Note that the output of this step is not the original text, but the one generated from the final set of claims. Therefore this text contains unfactual information. The generated text must follow the same logical flow as the ordering of the claims. The output text must be as similar as possible to the output of Step 2, unless the unfactual part. \\\\
\hline \\
\textbf{Output format:} Return the output in a JSON with the following format: \{ 'step\_1': List[str], 'step\_2': Tuple[str, str], 'step\_3': str, 'step\_4': str\}. The output must be a valid JSON, thus try to avoid special characters like ' and " inside the JSON values, unless you escape them with a \textbackslash. Do not include any marker for the altered claim inside the JSON values, e.g., \# this is the altered claim. Please do not provide any preamble to your response, just give me the JSON. \\\\
\hline
\end{tabular}
\caption{Prompt for the generation of data in \resource.}
\label{tab:table_prompt}
\end{table*}

\section{\resource\ \raisebox{-0.5ex}{\includegraphics[scale=0.05]{figures/oasis.png}}}
\label{sec:resource}
In this section, we outline the steps required to generate \resource. We start by selecting Wikipedia as our source of factual data due to its coverage of a wide range of topics and its frequent revisions, which help maintain accurate and up-to-date information. Moreover, to guarantee the quality of our data in terms of well-established and widely-referenced information, we retain the most popular English Wikipedia pages.\footnote{We select the 80k most visited pages in 2023.}  
Each page is then divided into passages of $K$ sentences using a sliding window with stride of $s$ sentences, forming our initial corpus.\footnote{In creating our resource, we set $K=5$ and $s=1$.}
Given a passage, as outlined in Fig.~\ref{fig:falsepedia}, we task an LLM\footnote{We used the GPT-4 API. More details in Appendix~\ref{app:llm_used}.} to: (i) extract a list of atomic claims (\textbf{Claim Extraction}, Sec.~\ref{sec:claim_extraction}); (ii) falsify one of the extracted claims (\textbf{Claim Falsification}, Sec.~\ref{sec:claim_falsification}); and (iii) generate a paraphrase of the original passage, grounded on the extracted claims, along with an unfactual version incorporating the information from the falsified claim (\textbf{Factual and Unfactual Text Generation}, Sec.~\ref{sec:text_generation}).

In the remainder of this section, for the sake of clarity, we describe the above-mentioned steps individually, but we anticipate that the step-specific outputs are obtained by means of a general, unified prompt containing the instructions for all the steps. The overall prompt is provided in Table \ref{tab:table_prompt}.

\subsection{Claim Extraction}
\label{sec:claim_extraction}
The first step in creating \resource~involves extracting claims from an input passage 
$t$. We randomly sample one passage from each Wikipedia page and then extract a list of claims from each of the passages (cf. Step 1 in Fig.~\ref{fig:falsepedia}).

We frame the claim extraction task as an end-to-end autoregressive generation problem. 
Let \( \mathcal{M} \) be our generative model. Given an input passage $t$, we task $\mathcal{M}$  to extract the claims using the prompt \( P_1(t) \) (cf. Step 1 in Table~\ref{tab:table_prompt}):
\begin{equation}
\label{eq:claim_extraction}
    \mathcal{M}(P_1(t))= (c_1,  \ldots, c_n) 
\end{equation}
where $(c_1, \ldots, c_n)$ represents the sequence of the generated claims. With the prompt $P_1(t)$, we aim at obtaining atomic\footnote{\citet{liu2023revisiting} defines a \textit{claim} as an Atomic Content Unit (ACU), that is, an elementary unit of information found in a text that does not require further subdivision for the purpose of reducing ambiguity.} and self-contained claims, i.e. elementary units of information that do not require additional context to be verified. Specifically, we explicitly require the model to adhere to such formal definition, and, additionally, constrain it to generate short texts and avoid the usage of pronouns as subjects.
For instance, given the \textbf{input passage}:
\begin{quote}
\textit{``The Amazon Rainforest, also known as Amazonia, is a moist broadleaf forest in the Amazon biome that covers most of the Amazon basin of South America. This region includes territory belonging to nine nations, with Brazil containing 60\% of the rainforest.''}
\end{quote}
the model $\mathcal{M}$ returns the following list of \textbf{claims}:
\begin{enumerate}
    \item The Amazon Rainforest is also known as Amazonia.
    \item It is a moist broadleaf forest in the Amazon biome.
    \item The Amazon Rainforest covers most of the Amazon basin of South America.
    \item The region includes territory belonging to nine nations.
    \item Brazil contains 60\% of the rainforest.
\end{enumerate}
Further examples of extracted claims can be found in Appendix~\ref{app:examples}.

\subsection{Claim Falsification}
\label{sec:claim_falsification}
With the aim of producing an unfactual version of the original text, we introduce a critical factual error into one of the extracted claims.
Formally, given the set of claims $C$ = $(c_1, \ldots, c_n)$ we task the model to falsify one of the claims\footnote{Our choice of falsifying only one claim per passage was intentional to increase the difficulty of the end-to-end factuality evaluation task. Identifying a text as “non-factual” when it contains multiple hallucinations is inherently easier than doing so when only a single, subtle falsehood is present.} as follows:

\vspace{-1em}
\begin{equation}
\label{eq:claim_falsification}
\begin{aligned}
\mathcal{M}(P_{2}(C))
&=  (c_{i}, \overline{c}_{i})
\end{aligned}
\end{equation}


\noindent where $P_{2}(C)$ is the prompt comprising the instructions for claim falsification, $\overline{c}_i$ the falsified claim and $c_i$ the corresponding factual one.
We ask the model to provide the factual claim as well, thus enabling the investigation of the model's behavior.

As outlined in Table ~\ref{tab:table_prompt} (Step 2), we instruct the model to falsify only one of the extracted claims by introducing a critical yet subtle error, which makes it potentially challenging to detect. 
Moreover, inspired by findings from previous works about the manual creation of Natural Language Inference (NLI) resources~\cite{parrish-etal-2021-putting-linguist, ocnli}, we designed the prompt with instructions to discourage the generation of naive contradicting instances, e.g., trivial negations of verbs. Continuing the example introduced in the previous section, given the extracted set of \textbf{claims}:  

\begin{enumerate}
    \item The Amazon Rainforest is also known as Amazonia.
    \item The Amazon Rainforest is a moist broadleaf forest in the Amazon biome.
    \item The Amazon Rainforest covers most of the Amazon basin of South America.
    \item The region includes territory belonging to nine nations.
    \item \colorbox{Verdetto}{Brazil contains 60\% of the rainforest.} ($c_i$)
\end{enumerate}

\noindent the model $\mathcal{M}$ produces the following \textbf{falsified claim}:

\begin{center}
\colorbox{Rossetto}{\parbox{0.8\linewidth}{The majority of the forest is contained within Peru.}}
($\overline{c_i}$)
\end{center}

In this example, the model replaces “Brazil” with “Peru”, another country partially covered by the Amazon rainforest, making the falsification subtle and contextually plausible. Unlike FEVER~\cite{thorne-etal-2018-fever}, which primarily focuses on controlled manipulations such as simple negations, our approach generates domain-specific substitutions that are deliberately more challenging to detect. 
Further examples of $\langle$factual, unfactual$\rangle$ pairs of claims can be found in Appendix \ref{app:examples}.


\subsection{Factual and Unfactual Text Generation}
\label{sec:text_generation}
Based on the claims extracted in the previous steps (cf. Sections~\ref{sec:claim_extraction} and \ref{sec:claim_falsification}), we now aim at generating pairs of $\langle$factual, unfactual$\rangle$ texts, which populate our resource for factuality evaluation, thus enabling the training and the benchmarking of factual reasoners. 

\paragraph{Factual text generation}
To make the factuality evaluation task more challenging, instead of using the original passages from Wikipedia as our factual texts, we leverage paraphrase generation. This approach produces texts that convey the same meaning as the original ones but with different surface forms, thereby making the verification task difficult for LLMs in both zero-shot settings -- as the original texts could have been seen during pretraining -- and RAG settings, which might retrieve the exact passages from Wikipedia.
Formally, given the set of extracted claims $C$, we task the model to generate a factual text $\mathcal{F}$ grounded on such claims:
\vspace{-1em}

\begin{equation}
\label{eq:paraphrase_generation}
\mathcal{M}(P_{3}(C)) =\mathcal{F}.
\end{equation}
\noindent  where $P_{3}(C)$ is the prompt with the instructions for obtaining a factual text through paraphrasing. 
As described in Table~\ref{tab:table_prompt} (Step 3) we explicitly require $\mathcal{M}$ to follow the sequence of extracted claims to encourage a full coverage of the facts expressed in the original text.
For instance, given the following \textbf{claims}:
\begin{enumerate}
    \item The Amazon Rainforest is also known as Amazonia.
    \item The Amazon Rainforest is a moist broadleaf forest in the Amazon biome.
    \item The Amazon Rainforest covers most of the Amazon basin of South America.
    \item The region includes territory belonging to nine nations.
    \item Brazil contains 60\% of the rainforest.
\end{enumerate}

\noindent the model $\mathcal{M}$ generates the following \textbf{factual text}: 
\begin{quote}
\textit{``Amazonia, widely known as the Amazon Rainforest, is a damp broadleaf forest located within the Amazon biome, covering a significant portion of the Amazon basin in South America. This vast region spans across nine countries, with Brazil housing 60\% of the rainforest.''}
\end{quote}

\noindent See Appendix~\ref{app:examples} for more examples of generated factual texts.

\paragraph{Unfactual text generation} Finally, the unfactual texts are generated through an analogous process, this time grounded on the set of claims that includes the unfactual one, namely, $\overline{C} =(c_{1}, \ldots, \overline{c}_{i}, \ldots, c_{n})$.
We obtain the unfactual text $\mathcal{U}$ with the generation process defined with the following:

\begin{equation}
\label{eq:paraphrase_generation}
\mathcal{M}(P_{4}(\overline{C}, \mathcal{F})) =\mathcal{U}
\end{equation}

\noindent where $P_{4}(\overline{C}, \mathcal{F})$ is the prompt containing the guidelines for unfactual text generation. In particular, as specified in Table~\ref{tab:table_prompt} (Step 4), we instruct $\mathcal{M}$ to produce a text identical to $\mathcal{F}$ except for the segment containing the factual error to ensure that the only confounding factor for the verification task is the unfactual portion of the text. This approach helps isolate the effect of the factual inaccuracy, preventing the model to introduce further inaccuracies. For example, given the \textbf{claims} in $\overline{C}$:
\begin{enumerate}
    \item The Amazon Rainforest is also known as Amazonia.
    \item The Amazon Rainforest is a moist broadleaf forest in the Amazon biome.
    \item The Amazon Rainforest covers most of the Amazon basin of South America.
    \item The region includes territory belonging to nine nations.
\item \colorbox{Rossetto}{\parbox{\linewidth}{The majority of the forest is contained within\\ Peru.}}\end{enumerate}

\noindent the model $\mathcal{M}$ generates the following \textbf{unfactual text}: 
\begin{quote}
\textit{``Amazonia, widely known as the Amazon Rainforest, is a damp broadleaf forest located within
the Amazon biome, covering a significant portion of the Amazon basin in South America. This
vast region spans across nine countries, and \colorbox{Rossetto}{\parbox{0.98\linewidth}{the majority of the forest is contained within Peru.}}
}
\end{quote}

As shown in this example, the falsification is seamlessly embedded within a factually accurate and natural-sounding passage. This introduces an additional layer of complexity compared to FEVER, where claims are presented in isolation for verification. Here, models must not only assess the factuality of individual statements but also distinguish between verifiable facts and misinformation carefully woven into coherent, credible narratives. Additional examples of unfactual texts can be found in Appendix~\ref{app:examples}. Finally, statistics about claim extraction, claim falsification, factual and unfactual text generation process can be found in Table~\ref{tab:falsepedia_statistics}.

\section{The \resource~benchmark}\label{sec:benchmark}

As a result of the steps described in Section \ref{sec:resource}, we obtained a large resource consisting of claims and texts (both factual and unfactual) that can be used to train end-to-end factuality evaluation systems. However, due to the automated nature of the proposed approach, it is crucial to both evaluate the quality of the produced data -- by accurately evaluating the individual steps of our pipeline -- and introduce a gold-standard benchmark for the task.

\begin{table}[t]
\centering
\begin{tabular}{l r}
\toprule
\multicolumn{2}{l}{\textbf{Claim Extraction}} \\
\# Pages & 81,275 \\
\# Passages & 81,275 \\
Avg. Tokens per Passage & 99.7 \\
\# Claims & 681,201 \\
Avg. Claims per Passage& 8.381 \\
Avg. Tokens per Claim & 8.6 \\
\midrule
\multicolumn{2}{l}{\textbf{Claim Falsification}} \\
\# Unfactual Claims & 81,275 \\
Avg. Tokens per Unfactual Claim & 9.0 \\
\midrule
\multicolumn{2}{l}{\textbf{Factual Text Generation}} \\
\# Factual Texts & 81,275 \\
Avg. Tokens per Factual Text & 82.9 \\
\midrule
\multicolumn{2}{l}{\textbf{Unfactual Text Generation}} \\
\# Unfactual Texts & 81,275 \\
Avg. Tokens per Unfactual Text & 86.5 \\
\bottomrule
\end{tabular}
\caption{Summary statistics for the creation of \resource.}
\label{tab:falsepedia_statistics}
\end{table}

\subsection{Human Evaluation}
\label{sec:quality_assessment}
To assess the quality of our dataset and enable a rigorous evaluation of our procedure, we asked $M=5$ expert linguists to validate a portion of $N=1,750$ instances for each task in our pipeline (cf. Sec. \ref{sec:resource} and Fig. \ref{fig:falsepedia}). Each annotator curated $(N/M)+K$ instances for each task with each of the $M$ subsets having an overlap of $K=100$ instances shared among all annotators. For the final benchmark, we resolve instances with disagreements through majority voting. We paid the annotators according to the standard salaries for their geographical location and provided them with task-specific guidelines, annotation examples, and a simple interface for each task. More details are provided in Appendix \ref{app:ann_guidelines}.

\begin{table}
\centering
\setlength{\tabcolsep}{5pt} 
\begin{tabular}{l|c|c}
\hline
\textbf{Task} & \textbf{Accuracy (\%)} & \textbf{Fleiss' $\kappa$} \\ \hline
Claim Extraction         & 96.78 & 0.81 \\   
Claim Falsification      & 98.55 & 0.84 \\ 
Factual Text Gen.        & 90.36 & 0.73 \\
Unfactual Text Gen.      & 89.20 & 0.72 \\ \hline
\end{tabular}
\caption{Performance of the chosen LLM $\mathcal{M}$ in the data generation process according to human evaluation (Accuracy), and the corresponding inter-annotator agreement (Fleiss' $\kappa$).}
\label{tab:human_evaluation}
\end{table}

\paragraph{Claim Extraction}
\label{par:claim_extraction}
For the claim extraction task, annotators received Wikipedia passages ($t_1, \ldots, t_N$), each accompanied by a list of claims extracted by the model 
$\mathcal{M}$ as described in Section \ref{sec:claim_extraction}. The annotators' task was to verify whether each claim was appropriately represented in the corresponding passage (i.e. with the same semantics) and assess their atomicity.\footnote{We chose to prioritize a precision-oriented evaluation for two key reasons: first, low coverage does not affect our proposed claim verification task (see Task 2, Section \ref{sec:gold}); and second, evaluating coverage would have required annotators to read the entire passage, making the annotation process more time-consuming and costly.}

We evaluated the LLM’s performance on this task by counting the human-annotated errors, yielding an accuracy of 96.78\%. Additionally, we measured inter-annotator agreement, resulting in a Fleiss’ $\kappa$ score of 0.81. These results underscore both the high quality of the generated $\langle$text, claims$\rangle$ pairs and the strong agreement among the annotators. 

Among the few errors produced by the LLM, we observe some occasional incorrect claims in the context of conditional clauses, where the model  interprets conditional or hypothetical statements as if they were factual claims. For instance, given the text: \textit{In contrast, \textbf{if} interest rates were the main motive for international investment, FDI \textbf{would} include many industries within fewer countries. [...]}, the following incorrect claims were extracted: \textit{Interest rates motivate international investment} and \textit{Interest rates lead to FDI in multiple industries}, thus misrepresenting the original text which, instead, indicates a hypothetical scenario.

\paragraph{Claim Falsification}
For this task, annotators received pairs of claims $\langle c_{i}, \overline{c_{i}}\rangle$ with $c_{i}$ being one of the original claims selected from ($c_1, \ldots, c_n$) and $\overline{c_{i}}$ the corresponding falsified claim produced by the model $\mathcal{M}$. The annotators' task was to verify whether each claim was appropriately falsified (i.e. with contradicting semantics). This required them to determine if $\overline{c_{i}}$ meaningfully diverged from $c_{i}$ in terms of content and truthfulness, effectively capturing the model's ability to produce altered, incorrect versions of the original claims. Again, the model achieved a very high accuracy ($98.55$\%). We measured a Fleiss' $\kappa $ score of $0.84$, showing up almost perfect agreement between the annotators.

In this case, one of the most frequent error category concerns instances where attempts at falsification manifest through minimal lexical variation, specifically by altering a single word. In these cases, such minor substitutions do not always yield a valid falsification. For example, consider the following claims: \textit{Michael Ausiello \textbf{authored} the exclusive piece} and	\textit{Michael Ausiello \textbf{wrote} the exclusive piece}. As we can see, despite the substitution of the verb, the semantic congruence between the two claims is maintained, rendering the falsification attempt ineffective. An additional instance of this type is represented by the claims: \textit{Washington, D.C. has \textbf{milder} winter weather than New York} and \textit{Washington, D.C. has \textbf{warmer} winter weather than New York}.

\paragraph{Factual and Unfactual Text
Generation} 
For these two tasks, we used a common format. Annotators received lists of original (or falsified) claims $C$ (or $\overline{C}$) and the associated factual (or unfactual) texts produced by the model $\mathcal{M}$. The annotators' task was to verify whether each claim was correctly represented in the generated text.
In the context of factual text generation, we additionally check whether the texts feature the same semantics as the claims but using a different wording. For the factual text generation step, we measured an accuracy of $90.36$\% and a Fleiss' $\kappa $ score of $0.73$. Similarly, for the unfactual text generation, we measured an accuracy of $89.2$\% and a Fleiss' $\kappa $ score of $0.72$. 

In the factual text generation task, we occasionally observe omissions of details present in the extracted claims. For instance, the month “May” is omitted in the factual rewriting of the claim “Russian President Yeltsin formed the Russian Armed Forces in May 1992”:

\begin{quote}
Originally, the Armed Forces of the Russian Socialist Federative Soviet Republic, also acknowledged as the Red Army, served both the Russian SFSR and Soviet Union. […] In 1992, Boris Yeltsin, the then Russian President, initiated the formation of the Russian Armed Forces, integrating a significant part of the Soviet Armed Forces.
\end{quote}

We also found similar omissions in some unfactual texts, where a factual claim extracted from the original passage is not included in the generated unfactual version. In both cases, we stress that these occasional omissions do not compromise the factuality labels of the generated texts. Our manual validation process confirmed that the omitted content was not critical for determining the factual status of the passage in all cases. However, when constructing our gold benchmark (Section 4.2), we prioritize precision by discarding all generated texts, both factual and unfactual, that any
annotator marks as containing an omission.





Overall, the reported detailed evaluations summarized in Table \ref{tab:human_evaluation} show the efficacy and robustness of the proposed methodology for producing training data for the task. 

\subsection{Gold Benchmark}\label{sec:gold}
In this section, we describe the construction of our  benchmark, along with the factuality-oriented tasks we propose.
Specifically, we exploit the human annotations (cf. Section \ref{sec:quality_assessment}) to construct a gold-standard benchmark for model evaluation.
To ensure the high quality of our data, we only retain the instances that were not marked as error by any of our annotators in any annotation stage (cf. Sec. \ref{sec:quality_assessment}).
We employ this data to propose the following two evaluation tasks, which we describe as follows.


\paragraph{{Task 1: End-to-End Factuality Evaluation}}
The first task is to determine whether a given text contains any factual inaccuracies. 
Formally, given an input passage \( t \), the model must output a binary label \( y \in \{\text{True}, \text{False}\} \), where \text{True} indicates that the text is factually accurate and \text{False} indicates the presence of factual inaccuracies.

For this setting, we rely only on factual and unfactual texts as input passages, and discard the original texts, as the latter might have already been seen during the pre-training of LLMs.
Specifically, to further ensure the high quality of our benchmark, we only retain the correct paraphrases that are generated from a valid set of claims (cf. \textit{Factual and Unfactual Text Generation} and \textit{Claim Extraction} in Sec. \ref{sec:quality_assessment}).
Concerning the valid unfactual texts, instead, we only keep the ones that are: i) generated, again, from a set of valid claims, and, ii) properly falsified and paraphrased (cf. \textit{Claim Falsification} and \textit{Factual and Unfactual Text Generation} in Sec. \ref{sec:quality_assessment}).
We then labeled all the resulting factual and unfactual texts with \textit{True}, and \textit{False}, respectively. 

In this setting, we aim at evaluating models on discerning true from fake texts (i.e., "Truth" from "Mirage"). This formulation enables the assessment of both plain LLMs and more complex RAG models. We deem this task to be particularly challenging as the falsification may involve even a single word occurring in one of the many claims featured in a text, in the spirit of recent works highlighting how LLMs struggle to deal with subtle nuances in a large input text \cite{kamradt2023, hsieh2024rulerwhatsrealcontext, laban2024summaryhaystackchallengelongcontext, wang2024multimodalneedlehaystackbenchmarking}

\paragraph{{Task 2: Evidence-based Claim Verification}}
In this setting, the task is to classify individual claims as factual or unfactual using a given evidence. This approach assumes that claims are already extracted from the text, simplifying the task by focusing on isolated statements rather than the entire text to verify. 
Formally, given an input claim \( c \) and a corresponding evidence passage \( e \), the model must output a binary label \( y \in \{\text{True}, \text{False}\} \), where \text{True} indicates that the claim is supported by the evidence and \text{False} indicates that the claim is not supported by the evidence. 


For this setting, we focus on the extracted claims and their corresponding unfactual version, and use the factual text as evidence.
We discard both the original and unfactual texts as the former might have already been seen during the pre-training of LLMs, while the latter contradicts real-world knowledge and, therefore, the internal knowledge of LLMs, possibly leading to unfair evaluations.

Additionally, to guarantee the high precision of our data, we focus on the claims that are both atomic and reflecting the same semantics of the original text (cf. \textit{Claim Extraction} Sec. \ref{sec:quality_assessment}). Then, we only keep the ones that have been appropriately falsified (cf. \textit{Claim Falsification} in Sec. \ref{sec:quality_assessment}), along with their unfactual counterparts. Finally, we apply the same quality checks described in Task 1 to retain only the valid factual texts.

At this stage, we classify the $\langle c_{i}, \mathcal{F} \rangle$ pairs with the label True, while we label $\langle \overline{c_{i}}, \mathcal{F}\rangle$ as False, with $c_{i}$ and $\overline{c_{i}}$ being the original claim and its falsified version, respectively.










\section{End-to-end Factuality Evaluation with \resource}\label{sec:approach}
In this section, we showcase how \resource~can be leveraged to build an end-to-end factuality evaluation system. In the spirit of \citet{min-etal-2023-factscore}, we decompose the task of evaluating the factuality of a given text into three simpler tasks, namely, Claim Extraction, Evidence Retrieval and Claim Verification. The process begins with extracting a set of atomic facts (cf. \textbf{Claim Extraction}, Sec. \ref{sec:approach_claim_extraction}) from the text to be verified. These extracted claims are then used to retrieve relevant evidence from a reliable knowledge base (cf. \textbf{Evidence Retrieval}, Sec. \ref{sec:approach_evidence_retrieval}). After this, the factual accuracy of each claim is evaluated by comparing it against the retrieved evidence (cf. \textbf{Claim Verification},  Sec.~\ref{sec:approach_claim_verification}). Finally, the results of these individual evaluations are aggregated to determine the overall factuality of the entire text.

\subsection{Claim Extraction}
\label{sec:approach_claim_extraction}
Our approach starts by extracting atomic claims from a given input text $t$. 
With the aim of training a claim extractor, we leverage \resource~to create a dataset of $\langle t$, $C\rangle$ tuples, where $t$ is an original text from Wikipedia and $C = (c_1, ..., c_n)$ the corresponding automatically-extracted claims by our chosen LLM $\mathcal{M}$ (Section 3.1). 
We then fine-tune a smaller sequence-to-sequence model $\mathcal{G}$ on this data, thus distilling the claim extraction capabilities of $\mathcal{M}$. 

We frame the training process as a text generation task; more formally, we fine-tune $\mathcal{G}$ to generate the claims autoregressively:

\begin{equation}
\label{eq:claim_gen}
P(y \mid t)=\prod_{k=1}^{|y|} P\left(y_k \mid y_{0: k-1}, t\right)
\end{equation}

\noindent where $y$ is the sequence obtained by concatenating the claims in $C$ and $y_k$ is a token in this sequence.
\subsection{Evidence Retrieval}
\label{sec:approach_evidence_retrieval}
At this stage, given the claims extracted by $\mathcal{G}$, we require a system capable of retrieving relevant passages from a knowledge corpus to serve as evidence to verify those claims. Again, we leverage \resource~to create a training dataset for our retriever; in particular, given each generic claim $c_j \in C$\ extracted from the original text $t$, we construct the following training pairs:
\begin{equation*}
\langle c_j, t \rangle, \langle c_j, \mathcal{F}\rangle,  \langle c_j, \mathcal{U}\rangle, \forall c_j \in C
\end{equation*}

\noindent where $\mathcal{U}$ and $\mathcal{F}$ are the generated factual and unfactual texts (cf. Section \ref{sec:text_generation}).

We then augment this set by pairing the factual and unfactual texts with the falsified claim $\overline{c_i}$ (cf. Section \ref{sec:claim_falsification}), thus obtaining the following additional training instances:
\begin{equation*}
\langle \overline{c_i}, t \rangle, \langle \overline{c_i}, \mathcal{F}\rangle, \langle \overline{c_i}, \mathcal{U}\rangle.
\end{equation*}

\noindent In this way, we include all possible pairs of $\langle$claim,  passage$\rangle$ in \resource~in our training set.
This strategy is aimed at increasing the generalization capabilities of our retriever: notably, given a claim, the retriever is trained to both provide the passages to support it along with the ones that are useful to contradict it.

Following the methodology outlined in Dense Passage Retrieval \citep[DPR]{karpukhin-etal-2020-dense}, we define our retriever $\mathcal{E}$ as a Transformer-based encoder, which produces dense representations of both claims and passages.
Starting from an input claim \( c \) and a knowledge corpus \( \mathcal{D} \), we use  $\mathcal{E}$ to compute a vector representation \( v_c \) for \( c \), and \( v_{p} \) for every passage \( \{p_1, p_2, \ldots, p_m\} \in\mathcal{D}\). Then, we use the dot product \( v_c \cdot v_{p} \) to rank all the passages in \( \mathcal{D} \) and, finally, extract the top \( k \) among these. 
The resulting $k$ passages form our evidence set \( R_k(c, \mathcal{D}) \) for \( c \).

Finally, we minimize the DPR loss $\mathcal{L}$ to train $\mathcal{E}$:

\vspace{-1em}
\begin{equation}
\mathcal{L} = - \sum_{i=1}^{N} \log \frac{e^{v_{c_i} \cdot v_{p_i^{+}}}}{e^{v_{c_i} \cdot v_{p_i^{+}}} + \sum_{j \neq i} e^{v_{c_i} \cdot v_{p_j^{-}}}}
\end{equation}
where \( N \) is the batch size, \( v_{c_i} \) is the vector representation of the \( i \)-th claim in the batch, \( v_{p_i^{+}} \) is the vector representation of the corresponding gold passage for the \( i \)-th claim, and \( v_{p_j^{-}} \) represents the vector representations of all the other passages in the batch, serving as in-batch negatives. This formulation ensures that the model learns to score the correct passage higher than the other ones within each batch, which has been shown to be an effective strategy for training retrieval models \cite{yih2011learning, gillick-etal-2019-learning}.

\subsection{Claim Verification}
\label{sec:approach_claim_verification}
The final step of our factuality evaluation methodology involves verifying each claim $c$ generated by our claim extractor from the text $t$, by comparing it against the corresponding passages \( R_k(c, \mathcal{D}) \) retrieved from our corpus.
Inspired by previous work on consistency  evaluation~\cite{zha2023alignscore, chen2023menli, scirè2024fenice}, we ground our verification approach on the NLI formulation. NLI is a task that determines the logical relationship between two texts: a \textit{premise} and a \textit{hypothesis}. Formally, given a premise $pre$ and a hypothesis $hyp$:
$\text{NLI}(pre, hyp) = Y\in \{\textsc{Ent}, \textsc{Neut}, \textsc{Contr}\}$, 
where $Y$ is a label indicating whether $pre$ entails (\textsc{Ent}), is neutral about (\textsc{Neut}), or contradicts (\textsc{Contr}) $hyp$.

\paragraph{Training a claim verifier on LLM-Oasis}
In this section, we show how \resource~can be utilized to train a model for the claim verification task. Complying with the NLI formulation, we require a strategy to assess whether each claim extracted from a text is entailed, contradicted, or neutral with respect to a set of the retrieved passages.
With this purpose, we construct a training dataset by deriving the following $\langle$claim, passage, label$\rangle$ triplets from \resource:
\begin{equation*}
    \langle c_j, t, \textsc{Ent} \rangle,
    \langle c_j, \mathcal{F}, \textsc{Ent}\rangle,  \forall c_j \in C
\end{equation*}
\noindent where $c_j \in C$ is a claim extracted by the LLM from the original text $t$ (cf. Section \ref{sec:claim_extraction}), and $\mathcal{F}$ and $\mathcal{U}$ are the factual and unfactual texts outlined in Section \ref{sec:text_generation}.

We expand our training dataset for NLI with the following triplets:
\begin{equation*}
    \begin{aligned}
        \langle \overline{c_{i}}, t,  \textsc{Contr} \rangle, 
        \langle \overline{c_{i}}, \mathcal{F}, \textsc{Contr}\rangle, \\
        \langle \overline{c_{i}}, \mathcal{U}, \textsc{Ent}\rangle, 
        \langle c_{i}, \mathcal{U}, \textsc{Contr}\rangle
    \end{aligned}
\end{equation*}\noindent where $\overline{c}_i$ is the falsified version of the extracted claim $c_i$ (cf. Sec. \ref{sec:claim_falsification}).\footnote{While edge cases exist where certain instances might be misclassified as contradictions, the low error rate observed in the claim falsification process (1.45\%) supports our decision to include these samples in the training dataset.}
To obtain a complete NLI dataset, we require a strategy to generate neutral triplets as well. To achieve this, we first pair each claim $c_j$ in $C$ (Section \ref{sec:claim_extraction}) with the passages $p_i$ of the Wikipedia page $W$ from where the original text $t$ was extracted. Then, we select the passage $p^*$ as the one that maximizes the neutrality probability when fed to an NLI model\footnote{We used a DeBERTa-v3-large model fine-tuned on several NLI datasets. For more information: \url{https://huggingface.co/MoritzLaurer/DeBERTa-v3-large-mnli-fever-anli-ling-wanli}} $\Psi$ along with $c_j$:

\[
p^*=\argmax_{p_i \in W} \mathbb{P}_{\Psi}(\textsc{Neut} \mid p_i, c_j)
\]

\noindent and augment our dataset with the neutral pairs \(\langle c, p^*, \textsc{Neut}\rangle\).
This approach increases the likelihood that the selected passages are semantically related to the claim, as they come from the same Wikipedia page, while still being neutral. This is preferable to randomly selecting neutral examples, as it tends to provide more meaningful contrasts.

Finally, we fine-tune a Cross-Encoder model on this data; as a result of this process, we obtained our claim verification model $\Phi$. More information about the training setup can be found in Section~\ref{sec:our_model_exp}.

\paragraph{Claim verification algorithm}
In Alg. \ref{alg:claim_verification} we outline how we leverage $\Phi$ to assess the factuality of a claim.
Our procedure takes as input a claim, a set of \textit{top-k} retrieved passages, and
a claim verification model. For each $\langle$passage, claim$\rangle$ pair we obtain a label $\hat{y}$ by applying $\Phi$:
\begin{equation}
\begin{aligned}
\hat{y} = \Phi( p_i, c) = \argmax_{y \in \{\textsc{Ent}, \textsc{Neut}, \textsc{Contr}\}} P(y \mid p_i, c)
\end{aligned}
\end{equation}

where $p_i$ is a retrieved passage and $c$ is a claim, which are fed to the NLI model $\Phi$ as the premise and hypothesis, respectively. 
As described in Alg. \ref{alg:claim_verification}, the algorithm proceeds by checking the output of this model for each passage in the ranking order. If $\Phi$ outputs $\textsc{Ent}$ for a passage, the claim is deemed verified (i.e., return True). Conversely, if $\Phi$ outputs $\textsc{Contr}$, the claim is deemed unfactual (i.e., return False). Finally, if $\Phi$ outputs $\textsc{Neut}$ for all passages, the claim is deemed verified (i.e., return True), as there is no contradicting evidence available.

In practice, given an input text $t$, we used our claim verifier to assign a factuality label to the claims generated by our claim extractor, using the passages returned by our retriever as evidences. 

The final factuality prediction for the text \( t \) is an aggregation of the claim-level factuality labels. Specifically, the text \( t \) is considered factual if all of its extracted claims are verified, unfactual otherwise.

\begin{algorithm}[t!]
\begin{algorithmic}[1]
\REQUIRE claim $c$, top-k retrieved passages $\{p_1, p_2, \ldots, p_k\}$, NLI model $\Phi$
\FOR{each passage $p_i$ in $\{p_1, p_2, \ldots, p_k\}$}
    \STATE $\hat{y} \leftarrow \Phi(c, p_i)$
    \IF{$\hat{y} == \textsc{Ent}$}
        \RETURN True 
    \ELSIF{$\hat{y} == \textsc{Contr}$}
        \RETURN False 
    \ENDIF
    \STATE \COMMENT{The output of the model is $\textsc{Neut}$, i.e., neutrality. Continue to the next passage}
\ENDFOR
\RETURN True \COMMENT{All NLI outputs are neutral, $c$ is deemed verified}

\caption{Algorithm for Claim Verification.}
\label{alg:claim_verification}

\end{algorithmic}
\end{algorithm}
\section{Experimental Setup}\label{sec:exp-setup}
In this section, we provide details about the models and data involved in our experiments. 
To train our components for the end-to-end factuality evalauation task, we leverage the synthetic data from \resource~(cf. Section~\ref{sec:resource}, Figure~\ref{fig:falsepedia}). Specifically, we randomly split the passages in an 80/20 proportion to build the train and validation datasets, respectively. When splitting, we ensure that all the claims, as well as the factual and unfactual text generated from the same passage, will end up in the same split.

We evaluate both our modular architecture (cf. Sec. \ref{sec:our_model_exp}) and several LLM-based baselines (cf. Sec. \ref{sec:baselines_exp_setup}), showing the effectiveness of our benchmark in challenging factuality evaluation systems. 
To assess their performance, we rely on the \resource~gold-standard benchmark (Section \ref{sec:gold}). 
Models are evaluated across the two proposed tasks (i.e. \textit{end-to-end verification} and \textit{evidence-based claim verification}), and we use balanced accuracy~\cite{5597285} as our evaluation metric. All fine-tuning experiments and inference for models up to 8B parameters are conducted on a single \textit{NVIDIA GeForce RTX 3090} GPU. For larger models, specifically Phi-4 and Llama-3.3-70B-Instruct, we utilize an HPC cluster node equipped with 4 \textit{NVIDIA A100} GPUs.

\subsection{Our model}
\label{sec:our_model_exp}
Here, we provide the training details for each module of our proposed solution for end-to-end factuality evaluation (cf. Sec. \ref{sec:approach}).

\paragraph{Claim extractor}
As described in Section~\ref{sec:approach_claim_extraction}, we build our claim extractor dataset with the $\langle$text, claims$\rangle$ tuples in the training split of \resource. We split the resulting dataset into $\sim$67k passage-claims pairs for training, and $\sim$4k passage-claims pairs for validation. Statistics about the claim extraction dataset can be found in Table \ref{tab:falsepedia_statistics}.

We fine-tune a T5$_{base}$~\citep{T5} model on this data to generate the sequence of claims given an input passage.
We train the model for a total of 1M steps, with Adafactor~\citep{adafactor} as optimizer with a learning rate of $1e^-5$.

Following \citet{scirè2024fenice}, we rely on the $\text{easiness}_{F1}$ metric for model  selection. Let $C$ represent the set of generated claims for a given text and $C^*$ the corresponding set of gold claims.  
To compute the $\text{easiness}_P$ score, as defined by \citet{zhang-bansal-2021-finding}, we first calculate the ROUGE-1\footnote{We consider ROUGE-1 to be a suitable basis for our easiness metric due to the high extractiveness of the claim extraction task.} score for each generated claim $c \in C$ by comparing it to every gold claim $c^*$ $\in C^*$, and then select the maximum score. The final $\text{easiness}_P$ score is obtained by averaging these maximum scores over all generated claims:

\begin{equation}
\label{eq:easiness_p}
\text{easiness}_P(C, C^*) = \frac{\sum_{c \in C}\max_{c^* \in C^*}\text{R1}(c, c^*)}{|C|}
\end{equation}

Similarly, we compute the $\text{easiness}_R$ score by selecting the maximum ROUGE-1 score for each gold claim $c^*$ with respect to all generated claims:

\begin{equation}
\label{eq:easiness_p}
\text{easiness}_R(C, C^*) = \frac{\sum_{c^* \in C^*}\max_{c \in C}\text{R1}(c, c^*)}{|C^*|}
\end{equation}

Finally, we combine $\text{easiness}_P$ and $\text{easiness}_R$ to calculate the $\text{easiness}_{F1}$ score, and select the model that achieves the highest $\text{easiness}_{F1}$ on our validation set.
\paragraph{Evidence Retriever} 
The training dataset of our retriever comprises $\sim$3.2M $\langle$claim-evidence$\rangle$ pairs. 
At validation/test time we construct the knowledge corpus with the original texts in our validation split and gold benchmark,  respectively.  
To make the evaluation more realistic and challenging, we expand the corpus with passages from the same Wikipedia page. This approach results in our corpus $\mathcal{D}$ comprising a total of 2.5M passages. 

We use the pre-trained Transformer-based architecture $E5_{base}$ \cite{wang2022text} as our encoder $\mathcal{E}$. To generate embeddings for both claims and passages, we apply mean pooling over the output of $\mathcal{E}$. The model is trained with a batch size of \num{20} input texts for \num{300000} steps, using AdamW \cite{loshchilov2018decoupled} as the optimizer. We employ a learning rate of \( 1 \cdot 10^{-6} \), with a \num{20}\% warm-up phase.

\paragraph{Claim Verifier}
As outlined in Section~\ref{sec:approach_claim_verification}, we formalize the claim verification task as an NLI problem and construct a dataset of $\sim$3.5M $\langle$premise, hypothesis, label$\rangle$ triplets from \resource. We devoted 3.2M instances for training our claim verification model and the remaining 300k for validation. We fine-tune DeBERTa-v3$_{large}$~\cite{he2021deberta} for a total of 1M steps on this data, using Adafactor.

\subsection{Evaluated LLMs}
\label{sec:baselines_exp_setup}
We provide a comprehensive evaluation of a set of LLMs on the \resource~benchmark. 
We evaluate a closed‐source model from the GPT family—specifically, GPT‑4o~\cite{openai2024gpt4}—alongside open-weight LLMs such as Qwen‑2.5~\cite{qwen2025qwen25technicalreport}, Llama 3~\cite{grattafiori2024llama3herdmodels}, Mistral~\cite{jiang2023mistral7b}, Phi‑4~\cite{abdin2024phi4technicalreport}, and Phi‑4‑mini~\cite{microsoft2025phi4minitechnicalreportcompact}, as well as Falcon‑Mamba~\cite{zuo2024falconmambacompetitiveattentionfree}, which serves as a representative of non‐Transformer‐based architectures. These models were selected for their widespread use in the literature and their demonstrated high performance on standard evaluation benchmarks. By analyzing systems with parameter counts ranging from 4B to 70B, we can assess how different architectural approaches perform on factuality evaluation tasks.

\subsection{Evaluation settings}
Following standard practice in LLM evaluation, we assess model performance across multiple prompting strategies. Specifically, we consider four settings: Zero-Shot (ZS), Few-Shot (FS), Explain-Then-Answer (EXP), and Retrieval-Augmented Generation (RAG).

\paragraph{Zero-Shot (ZS)}
In this setting, the LLMs are prompted with the instructions and the input without any additional guidance. This setting serves as a baseline to assess the model’s inherent ability to evaluate factuality.

\paragraph{Few-Shot (FS)}
To guide the model in performing the task, we employ a few-shot learning approach by including a set of 5 manually-labeled held-out examples within the prompt.

\paragraph{Explain-then-Answer}
In this setting, the model is required to generate an explanation before providing a factuality label. This structured response format encourages the model to engage in explicit reasoning, potentially making its decision process more interpretable and accurate. 
\begin{figure*}[t]
    \centering
    \includegraphics[width=1\textwidth, trim=0 0 0 25, clip]{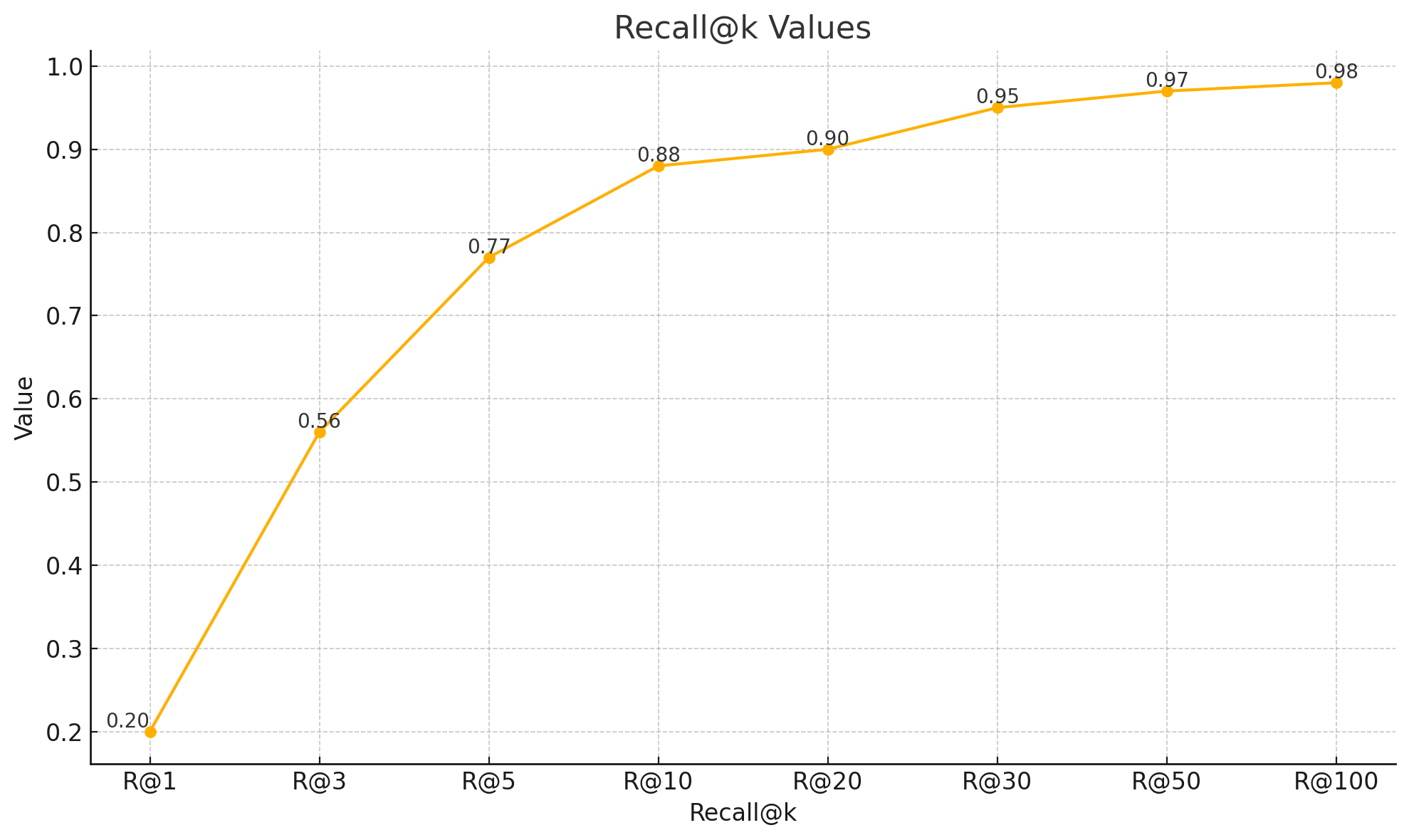}
    \caption{Recall@k performance of the $E5_{base}$ model at different values of $k$.}
    \label{fig:recall-k-results}
\end{figure*}
\paragraph{Retrieval Augmented Generation (RAG)}
As part of the end-to-end task evaluation, we ablate the impact of providing the LLMs with external knowledge, that is, in the RAG setting. To experiment with this, we include the top-$K$ passages\footnote{We selected $K$=30 based on the analysis of our retriever's performance at different values of $K$ (cf. Sec. \ref{sec:e2e_fact_experiments}) conducted on the validation set.} returned by our retriever (cf. Section~\ref{sec:approach_evidence_retrieval}) in the prompts. We extend the input by appending the retrieved passages after the text to be verified and a separator.\\\\
All the prompts used in the various settings can be found in Appendix~\ref{app:prompt_e2e}.

\section{Results}\label{sec:results}
\subsection{Task 1: End-to-End Factuality Evaluation}
\label{sec:e2e_fact_experiments}
In this section we present the results obtained in the end-to-end factuality evaluation task (cf. Section \ref{sec:gold}). First of all, we examine the performance of the evidence retrieval module, as this component supplies the external knowledge that is fed to the claim verifier. The performance of the end-to-end process depends on the quality of the retrieved evidence. This step also establishes an upper bound on the external knowledge integration, directly impacting the subsequent evaluation results.

\paragraph{Evidence Retriever}
We evaluate the performance of the evidence retrieval module using the Recall at $k$ (R@k) metric, which quantifies the proportion of relevant documents retrieved in the top $k$ results. Formally, it is defined as:

\begin{equation}
\text{R@k} = \frac{|\{\text{relevant } D\} \cap \{\text{top } k \text{ retrieved } D\}|}{|\{\text{relevant } D\}|}
\end{equation}

This metric allows us to assess the ability of our retriever to identify relevant passages for factuality verification within the top-$k$ ranked results. Higher values of $k$ generally yield higher recall, as more documents are considered, but also introduce the risk of increasing irrelevant retrievals.

For our experiments, we evaluated different values of $k$ (as shown in Figure \ref{fig:recall-k-results}) and ultimately selected for all the subsequent experiments $k=30$ as it provided a balance between performance and efficiency. The fine-tuned $E5_{base}$ model achieved a Recall@30 (R@30) of 0.95. This is a significant improvement compared to the same model without fine-tuning, which only achieved an R@30 of 0.52. The fine-tuning process over 3.2M passages proved crucial for this performance gain. We remark that R@K represents an upper bound of our factuality evaluation performance when external knowledge is integrated into the verification process. Further analysis and details can be found in Appendix \ref{app:retriever}.

\begin{table*}[t]
\centering
\setlength{\tabcolsep}{4pt}
\begin{tabular}{llcccc}
\toprule
\textbf{Model} & \textbf{Size} & \textbf{ZS} & \textbf{ZS+EX} & \textbf{FS} & \textbf{FS+EX} \\
\midrule
Phi-4-mini-instruct & 4B & 54.4{\scriptsize±0.4} & 52.6{\scriptsize±0.5} & 53.3{\scriptsize±0.4} & 52.3{\scriptsize±0.5} \\
Falcon3-Mamba-7B-Instruct & 7B & 55.3{\scriptsize±0.4} & 52.3{\scriptsize±0.5} & 53.0{\scriptsize±0.4} & 54.1{\scriptsize±0.4} \\
Mistral-7B-Instruct-v0.3 & 7B & 51.2{\scriptsize±0.5} & 54.6{\scriptsize±0.4} & 53.4{\scriptsize±0.4} & 56.2{\scriptsize±0.5} \\
Qwen2.5-7B-Instruct & 7B & 55.8{\scriptsize±0.4} & 52.7{\scriptsize±0.5} & 57.2{\scriptsize±0.4} & \underline{57.3}{\scriptsize±0.4} \\
Llama-3.1-8B-Instruct & 8B & 53.6{\scriptsize±0.3} & 54.9{\scriptsize±0.4} & 55.0{\scriptsize±0.4} & 55.5{\scriptsize±0.5} \\
Phi-4 & 14B & \underline{57.2}{\scriptsize±0.3} & \textbf{57.9}{\scriptsize±0.4} & \underline{57.6}{\scriptsize±0.4} & 57.0{\scriptsize±0.4} \\
Llama-3.3-70B-Instruct & 70B & \textbf{59.2}{\scriptsize±0.4} & \textbf{57.5}{\scriptsize±0.4} & \textbf{61.7}{\scriptsize±0.4} & \textbf{60.0}{\scriptsize±0.4} \\
\bottomrule
\end{tabular}
\caption{Balanced accuracy (\%) on the gold benchmark of \resource{} for end-to-end factuality evaluation. We report results of LLMs across different settings: Zero-Shot (ZS), Few-Shot (FS), and their respective Explain-the-Answer variants (ZS+EX, FS+EX). “Size” denotes the number of parameters in billions (B). Results are averaged over five runs with different seeds; standard deviations are reported as subscripts.}
\label{tab:e2e_all_models}
\end{table*}

\begin{table*}[t]
    \centering
    \begin{tabular}{lc|cc}
    \toprule
    \textbf{Model} & \textbf{Size} & \multicolumn{2}{c}{\textbf{B-Accuracy (\%)}} \\
    \cmidrule(lr){3-4}
    & & \textbf{ZS} & \textbf{RAG} \\
    \midrule
    Phi-4-mini-instruct & 4B & 54.4\tiny{$\pm0.4$} & 56.3\tiny{$\pm0.4$} \\
    Falcon3-Mamba-7B-Instruct & 7B & 55.3\tiny{$\pm0.4$} & 51.0\tiny{$\pm0.6$} \\
    Mistral-7B-Instruct-v0.3 & 7B & 51.2\tiny{$\pm0.5$} & 50.5\tiny{$\pm0.5$} \\
    Qwen2.5-7B-Instruct & 7B & 55.8\tiny{$\pm0.4$} & 54.7\tiny{$\pm0.4$} \\
    Llama-3.1-8B-Instruct & 8B & 53.6\tiny{$\pm0.3$} & 54.1\tiny{$\pm0.4$} \\
    Phi-4 & ~14B & 57.2\tiny{$\pm0.3$} & 57.6\tiny{$\pm0.4$} \\
    Llama-3.3-70B-Instruct & ~70B & 59.2\tiny{$\pm0.4$} & 58.8\tiny{$\pm0.4$} \\
    GPT-4o & ~~N/A & \textbf{60.8}~~~~~~& \underline{68.0}~~~~~ \\
    \midrule
    \textbf{Our Model (Fine-tuned)} & \textbf{1B} & - & ~~\textbf{69.24\tiny{$\pm0.4$}} \\
    \bottomrule
    \end{tabular}
    \caption{Balanced accuracy (B-Accuracy) on the gold benchmark of \resource{} for end-to-end factuality evaluation. We compare different models in Zero-Shot (ZS) and RAG settings. “Size” denotes the number of parameters in billions (B). Results for open-weight models are averaged over five runs with different random seeds, and standard deviations are reported as subscripts.}
    \label{tab:ours_and_closed}
\end{table*}

\begin{figure*}[t!]
    \centering
    \includegraphics[scale=0.55]{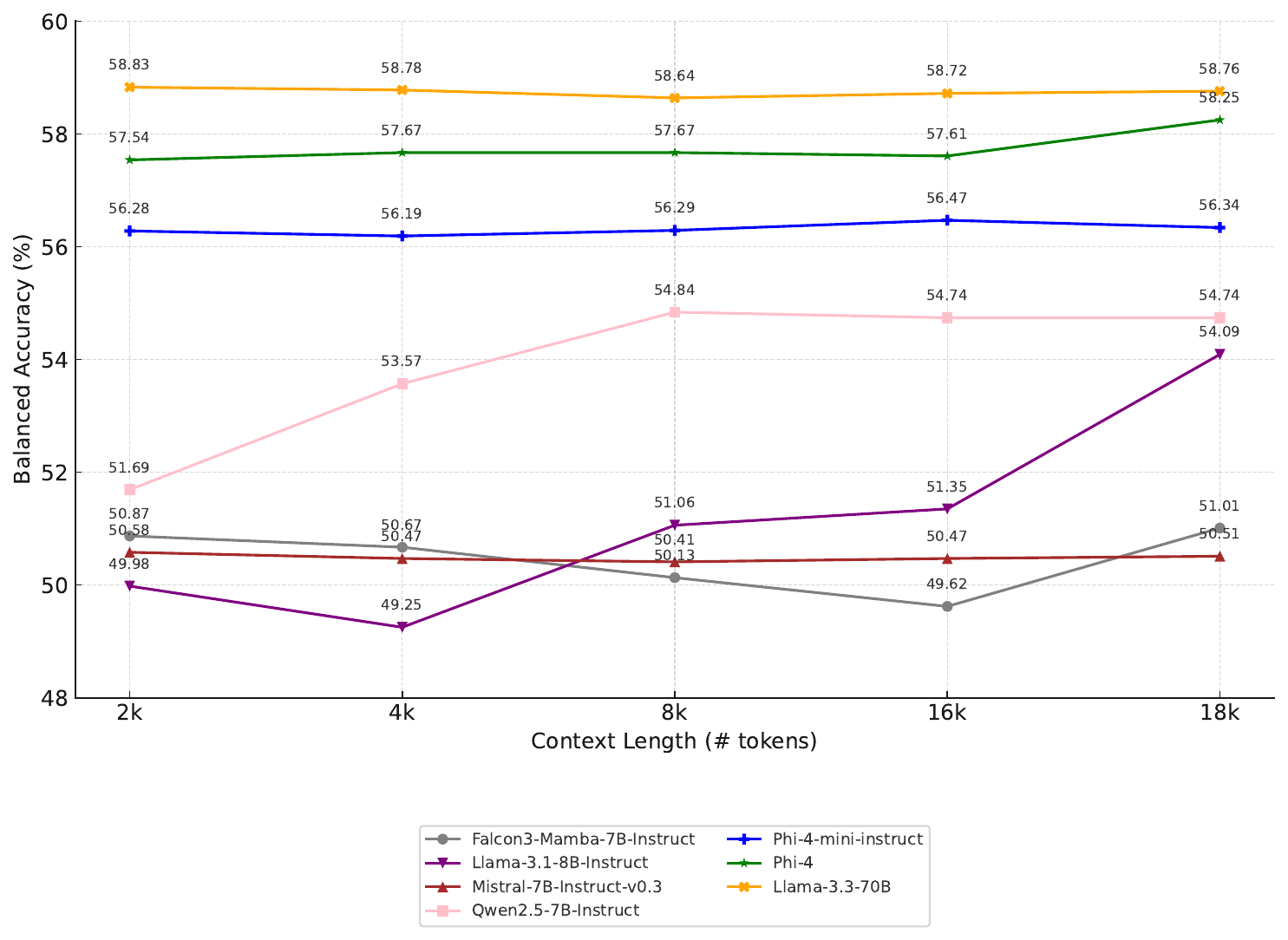}
    \caption{Balanced Accuracy (\%) of different models in the RAG setting with increasing context length. Each model is evaluated on inputs of 2k, 4k, 8k, 16k, and 18k tokens, which marks the length of the longest prompt instantiated in our evaluation.}
    \label{fig:rag_context_length}
\end{figure*}

\paragraph{End-to-End Factuality Evaluation}
The results of the evaluated LLMs for the end-to-end factuality evaluation task are shown in Table  \ref{tab:e2e_all_models}. We conduct evaluations in Zero-Shot (ZS) and Few-Shot (FS) prompting, with each setting also assessed using the Explain-Then-Answer (EX) approach.
As shown, the balanced accuracy scores of all evaluated LLMs remain low, often only marginally surpassing the random baseline. The top-performing model across all configurations is, unsurprisingly, the largest one—Llama-3.3-70B-Instruct—yet it reaches a maximum of just 61.7\% accuracy in the FS setting. This outcome highlights that our benchmark is extremely challenging even for state-of-the-art LLMs. The core difficulty lies in the nature of the task: models must assess the factuality of a text containing a subtle falsification seamlessly embedded within an otherwise factual context (cf. Section~\ref{sec:claim_falsification}). Moreover, this shows that despite having been likely exposed to the entire Wikipedia during the pretraining phase, the evaluated models still struggle to assign correct factuality labels.

In Table~\ref{tab:ours_and_closed} we report the performance of our approach (cf. Section \ref{sec:approach}) compared to all the other evaluated models in ZS and RAG settings. 
Results show that our pipeline-based approach using small language models (cf. Section~\ref{sec:approach}) achieves the highest balanced accuracy (69.24\%), significantly surpassing all evaluated LLMs— with only GPT-4o, in the RAG setting, approaching comparable performance. Although this outcome can be partly attributed to the fine-tuning of our model components on \resource, it remains notable given that our system operates with significantly fewer parameters than its counterparts. We argue that this advantage stems not only from the quality of the training data but also from the modular design of our approach, which decomposes the factuality evaluation task into simpler subtasks. This structure allows small models to effectively handle each task, leading to overall performance on par with, or exceeding, that of much larger LMs.
However, the fact that the best-performing system achieves a score of $\sim0.70$ further highlights the complexity of the proposed benchmark and paves the way for future studies on factuality evaluation.

The results from Table~\ref{tab:ours_and_closed} also suggest that incorporating retrieved external knowledge not only fails to consistently boost performance but often results in degradation compared to the Zero-Shot (ZS) setting, with most LLMs—including Llama-70B—struggling to effectively leverage the retrieved evidence. To further investigate this, we assess whether increased context length confounds models in this setting, we evaluate LLMs while truncating their input at different lengths. Figure~\ref{fig:rag_context_length} reports balanced accuracy across varying input lengths, with 18k tokens marking the length of the longest prompt instantiated in our evaluation. Across most models, performance remains relatively stable as input size increases. Notably, Qwen2.5-7B-Instruct and Llama-3.1-8B-Instruct exhibit moderate upward trends, suggesting some benefit from extended context.
These results suggest that context length alone does not hamper model performance.
Importantly, at 18k tokens, we guarantee that in 95\% of cases, the passage required to verify the text is present in the retrieved evidence (cf. Figure~\ref{fig:recall-k-results}). This setup allows to decouple evidence availability from reasoning: failures are most likely due to difficulties in exploiting the available information than to missing context. GPT-4o stands out in this regard, showing a +8 point gain in the RAG setting over ZS (cf. Table\ref{tab:ours_and_closed}), suggesting a more effective reasoning capability over retrieved evidence compared to the evaluated open-weights models.

\begin{table*}[t]
    \centering
    \begin{tabular}{lc|cccc}
    \toprule
    \textbf{Model} & \textbf{Size} & \textbf{ZS} & \textbf{ZS+EX} & \textbf{FS} & \textbf{FS+EX} \\
    \midrule
    Phi-4-mini-instruct & 4B & 70.50\tiny{$\pm0.5$} & 84.06\tiny{$\pm0.6$} & 56.56\tiny{$\pm0.6$} & 54.78\tiny{$\pm0.5$} \\
    
    Falcon3-Mamba-7B-Instruct & 7B & 63.45\tiny{$\pm0.7$} & 74.98\tiny{$\pm0.5$} & 67.97\tiny{$\pm0.6$} & 69.89\tiny{$\pm0.6$} \\
    Mistral-7B-Instruct-v0.3 & 7B & 72.87\tiny{$\pm0.5$} & 80.67\tiny{$\pm0.5$} & 73.40\tiny{$\pm0.6$} & 77.81\tiny{$\pm0.6$} \\
    Qwen2.5-7B-Instruct & 7B & 84.85\tiny{$\pm0.6$} & 87.11\tiny{$\pm0.5$} & 84.17\tiny{$\pm0.7$} & 84.25\tiny{$\pm0.6$} \\
    Llama-3.1-8B-Instruct & 8B & 78.56\tiny{$\pm0.6$} & 86.55\tiny{$\pm0.7$} & 76.94\tiny{$\pm0.5$} & 76.53\tiny{$\pm0.6$} \\
    
    Phi-4 & ~14B & 84.14\tiny{$\pm0.4$} & 92.69\tiny{$\pm0.5$} & 87.57\tiny{$\pm0.5$} & 88.32\tiny{$\pm0.4$} \\
    Llama-3.3-70B-Instruct & ~70B & \textbf{91.79}\tiny{$\pm0.4$} & \textbf{93.97}\tiny{$\pm0.3$} & \textbf{92.77}\tiny{$\pm0.3$} & \textbf{94.08}\tiny{$\pm0.4$} \\
    GPT-4o & ~~N/A & \underline{89.49}~~~~~ & \textbf{93.93}~~~~~ & \underline{88.24}~~~~~ & \underline{90.82}~~~~~ \\
    \bottomrule
    \end{tabular}
    \caption{Balanced accuracy (\%) on the gold benchmark of \resource{} for evidence-based claim verification. We compare different models across evaluation settings: Zero-Shot (ZS), Zero-Shot with Explanation (ZS+EX), Few-Shot (FS), and Few-Shot with Explanation (FS+EX). “Size” denotes the number of parameters in billions (B). Results for open-weight models are averaged over five runs with different random seeds, and standard deviations are reported as subscripts.}
    \label{tab:claim_verification_with_evidence}
\end{table*}

\begin{table*}[t!]
    \centering
    \begin{tabular}{lc|c}
    \toprule
    \textbf{Model} & \textbf{Size} & \textbf{B-Accuracy (\%)} \\
    \midrule
    \textbf{Our Model (Fine-tuned)} & ~~~0.4B & 93.30\tiny{$\pm0.4$} \\
    \bottomrule
    \end{tabular}
    \caption{Balanced accuracy (\%) of our claim verification model fine-tuned on \resource{}, evaluated on the gold benchmark. Results are averaged over five runs with different random seeds; standard deviation is reported as a subscript.}
    \label{tab:claim_verifier}
\end{table*}

\subsection{Task 2: Evidence-based Claim Verification}
\label{sec:evidence_eval_experiments}
In this section, we present the results for the second task we aim to evaluate with our benchmark (see Section \ref{sec:gold}), i.e., evidence-based claim verification.
We first evaluate the performance of the LLMs across the studied prompt settings (ZS, ZS+EX, FS, and FS+EX). The outcomes of these experiments are presented in Table~\ref{tab:claim_verification_with_evidence}. 
Notably, all systems  
achieve higher performance compared to the previous setting (e.g., our system goes from 69.24 in the end-to-end task to 93.30 in this task).  We attribute this to three main factors. First, this task is a simpler instance of the previous one, namely, the model is required to verify a single claim rather than a passage. 
Second, the system is provided with the exact evidence needed to verify the claim while, in the end-to-end formulation, each model relies on several passages returned by the retriever, hence possibly introducing noise in the process. Finally, the end-to-end verification implies reading and reasoning on a huge context (4k tokens on average) rather than the limited one (100 tokens on average) of this task. 

To assess the effectiveness of a specialized model trained directly on our resource, we evaluate our claim verifier (cf. Section~\ref{sec:approach_claim_verification}), thereby excluding the claim extraction and retrieval components from our pipeline. The results, shown in Table~\ref{tab:claim_verifier}, indicate that our lightweight finetuned system (0.4B) obtains a very high balanced accuracy (93.30\%). This provides further evidence that fine-tuning on high-quality task-specific data can enable a small model to rival or even outperform much larger LLMs in factuality evaluation tasks.


\section{Conclusion and Future Work}
In this paper, we introduce \resource, a large-scale  resource for end-to-end factuality evaluation obtained by extracting and falsifying information from Wikipedia. Specifically, as outlined in Figure \ref{fig:falsepedia}, given a text from Wikipedia, we extract a set of factual and unfactual claims, with the latter obtained by falsifying one of the facts expressed in the original text. Starting from these sets, we design two \textit{claims2text} tasks and generate a factual text, which is a paraphrase of the original one, and its unfactual counterpart, featuring the falsified claim. 
This resulted in \numInstances~$\langle \text{factual}, \text{unfactual} \rangle$ pairs that are suitable for training factuality evaluation systems, making  \resource~the largest resource for this task. Contrarily to previous works in this domain, such as FEVER, which is focused on the simpler task of claim verification, our resource is the first enabling the training of end-to-end factuality evaluation systems, i.e., approaches that are able to assess the factuality of generic text in natural language.

We additionally devise a human annotation process to create a gold standard for benchmarking factuality evaluators and to validate the quality of the proposed data creation pipeline. \resource~enables two challenging tasks: \textit{end-to-end factuality evaluation}, which tests the ability of models to verify factual accuracy in  raw texts in natural language, and \textit{evidence-based claim verification}, which focuses on assessing individual claims against provided evidence. 

Our experiments reveal that open weights LLMs, such as Phi-4 and Llama 3, fall short in the end-to-end task, only marginally surpassing the random baseline. 
In the same setting, even GPT-4o faces significant challenges, in both zero-shot and RAG settings, i.e., when provided with supporting evidence from Wikipedia, only achieving 
60\% and 68\% of accuracy, respectively. This underscores the difficulty of the proposed benchmark and its potential to drive progress in factuality evaluation. Furthermore, thanks to \resource, we designed a novel baseline for end-to-end factuality evaluation, which consists of a pipeline of smaller, specialized models trained on three subtasks, namely, claim extraction, evidence retrieval and claim verification. Our approach demonstrated competitive or even superior performance to GPT-4o, showcasing the potential of smaller LMs fine-tuned on specific data for factuality evaluation.

Looking forward, we plan to expand \resource~to incorporate data from diverse domains and multiple languages, enhancing its utility and applicability. With the aim of fostering research in factuality evaluation, we release our resource at \url{https://github.com/Babelscape/LLM-Oasis}.

\section{Challenges and Discussion}
In this section, we reflect on a number of relevant aspects emerging from the design and construction of our benchmark, including open challenges, modeling decisions, and future directions for improving factuality evaluation.

\paragraph{Quality of the silver data}
Even if we manually-validate a subset of the data, our resource is LLM-generated. The utilized model, namely GPT-4, achieved very high performance in the various generation tasks (cf. Sec.~\ref{sec:benchmark}), but the introduced errors, even if they are few, may affect the quality of the training dataset. For this reason, we suggest to leverage the automatically-generated portion of our resource for developing systems, rather than benchmarking, for which we direct to our gold standard benchmark.

\paragraph{Multi-step prompting}
A potential limitation of our dataset generation process is the adoption of a unified prompt in a single API call to GPT-4, rather than employing a multi-step prompting strategy. While multi-step prompting could, in principle, improve the performance of individual data generation stages (e.g., claim extraction, falsification, and paraphrasing), we opted for a single-prompt approach primarily due to budget constraints. Using GPT-4, which is a paid model, a multi-step strategy would have significantly increased the number of API calls, as each step would require re-sending the entire Wikipedia passage. This would have resulted in approximately four times the cost, due to higher input token usage across multiple calls. However, we qualitatively compared both approaches and observed no substantial improvement in the quality of the generated outputs. Therefore, we adopted the single-prompt strategy as a more efficient and cost-effective solution, without compromising the integrity of the generated data.

\paragraph{Reliance on Wikipedia}
Additionally, \resource~is limited to Wikipedia as the source of factual information. This restricts the diversity of the dataset and may not other kinds of texts, such as scientific articles or news. 
We also note that our end-to-end evaluation task may require periodic updates as Wikipedia evolves: while unfactual texts generally remain valid over time, factual ones could become outdated. To maintain long-term relevance, future iterations of the dataset will incorporate updated Wikipedia dumps, ensuring that the benchmark remains challenging as LLMs get exposed to updated knowledge.

\paragraph{Rarity of the falsified facts}
We acknowledge that rare or less frequent facts—typically referred to as the long tail—represent a known challenge for factuality evaluation systems, including ours. While our modular pipeline does not rely on domain-specific priors and is, in principle, extendable to less frequent content, performance may degrade if relevant knowledge is underrepresented in the training data of each component.

That said, we argue that LLM-Oasis already provides a challenging setting, even without explicitly targeting long-tail phenomena. First, we note that state-of-the-art open-weight LLMs, even in Few-Shot settings, do not surpass 61.7 \% accuracy on our benchmark—highlighting the inherent difficulty of the task. This suggests that factuality evaluation remains an open problem even when models are tested on commonly known entities.

Second, although we constructed our dataset from the top 100k most-viewed Wikipedia pages to ensure quality and consistency, popularity at the page level does not imply that all facts within that page are well-known or frequently mentioned elsewhere. Indeed, popular entries often contain historical nuances, or lesser-known anecdotes that are less likely to be memorized or represented in LLMs’ training data.

Finally, assessing factual rarity is not trivial. The frequency of a fact is hard to quantify reliably, as it can be expressed in many ways across Wikipedia. For this reason, we believe that increasing the dataset’s coverage of long-tail content remains a valuable future direction, but our current benchmark already captures a wide factual spectrum and exposes significant limitations in existing systems.

\paragraph{Multilinguality} Finally, our analysis and experiments are limited to English-only data, which constrains the applicability of our findings to other languages. 

In future iterations, we plan to extend LLM-
Oasis to include a broader range of domains and languages, in order to better support multilingual and cross-domain factuality evaluation.

\bibliography{main}

\appendix
\section{Examples of Generated Texts}
\label{app:examples}

We present several examples derived from our dataset to show the model's capability of generating both factual and unfactual texts. These examples show how our pipeline produces paraphrased versions of original texts and introduces subtle yet critical factual inaccuracies.

\subsection{Example 1}

\textbf{Original Text:}

\textit{Albert Einstein was a German-born theoretical physicist who developed the theory of relativity, one of the two pillars of modern physics. His work is also known for its influence on the philosophy of science. Einstein is best known for his mass–energy equivalence formula $E = mc^2 $, which has been dubbed “the world’s most famous equation”.}

\textbf{Extracted Claims:}
\begin{enumerate}
\item Albert Einstein was a German-born theoretical physicist.
\item He developed the theory of relativity.
\item The theory of relativity is one of the two pillars of modern physics.
\item Einstein’s work influenced the philosophy of science.
\item He is best known for his mass–energy equivalence formula $E = mc^2 $.
\item The formula $E = mc^2 $ is dubbed “the world’s most famous equation”.
\end{enumerate}

\textbf{Factual Text:}

\textit{Albert Einstein, originally from Germany, was a theoretical physicist who formulated the theory of relativity, a cornerstone of modern physics. His contributions significantly impacted the philosophy of science. The mass–energy equivalence equation  $E = mc^2$, which he is most famous for, is often called “the world’s most famous equation”.}

\textbf{Falsified Claim:}

\textit{He developed the theory of quantum mechanics.}

\textbf{Unfactual Text:}

\textit{Albert Einstein, originally from Germany, was a theoretical physicist who formulated the theory of quantum mechanics, a cornerstone of modern physics. His contributions significantly impacted the philosophy of science. The mass–energy equivalence equation $E = mc^2 $ , which he is most famous for, is often called “the world’s most famous equation”.
}
\subsection{Example 2}

\textbf{Original Text:}

\textit{The Amazon Rainforest, also known as Amazonia, is a moist broadleaf forest in the Amazon biome that covers most of the Amazon basin of South America. This region includes territory belonging to nine nations. The majority of the forest is contained within Brazil, with 60\% of the rainforest.}

\textbf{Extracted Claims:}
\begin{enumerate}
\item The Amazon Rainforest is also known as Amazonia.
\item It is a moist broadleaf forest in the Amazon biome.
\item The Amazon Rainforest covers most of the Amazon basin of South America.
\item The region includes territory belonging to nine nations.
\item The majority of the forest is contained within Brazil.
\item Brazil contains 60\% of the rainforest.
\end{enumerate}

\textbf{Factual Text:}

\textit{Amazonia, widely known as the Amazon Rainforest, is a damp broadleaf forest located within the Amazon biome, covering a significant portion of the Amazon basin in South America. This vast region spans across nine countries, with Brazil housing 60\% of the rainforest.
}

\textbf{Falsified Claim:}

\textit{The majority of the forest is contained within Peru.}

\textbf{Unfactual Text:}

\textit{Amazonia, widely known as the Amazon Rainforest, is a damp broadleaf forest located within the Amazon biome, covering a significant portion of the Amazon basin in South America. This vast region spans across nine countries, with Peru housing 60\% of the rainforest}

These examples show the effectiveness of the model in creating pairs of factual and unfactual texts. The alterations are subtle, ensuring that the generated unfactual texts are challenging for both human annotators and automated systems to detect, thus providing a robust testbed for evaluating the factual accuracy of LLM-generated content.

\section{Prompts for End-to-End Factuality Evaluation}
\label{app:prompt_e2e}
To accomplish the task of end-to-end factuality evaluation, we employ different strategies depending on the language model being used. For models like Llama, which supports a system prompt, we set specific instructions as the system message. For models like Mistral, which do not support a system prompt, we include the instructions at the beginning of the text. In our experiments, we set the temperature to 0.0 to guarantee consistency of the results across different runs.

The prompts used for factuality evaluation in Zero-Shot and RAG are displayed in Table \ref{tab:zs_prompt} and \ref{tab:rag_prompt}. 
In the latter setting, we prompted all the LLMs with the same pieces of evidence retrieved and used by our NLI module (cf. Sec. \ref{sec:e2e_fact_experiments}).
Concerning the Explain-then-Answer paradigm, we expand the set of instructions with the following recommendation:
\\\\
Motivate your response with an explanation and then reply with "Factual" or "Not Factual"\\
Output format:\\
\#\# EXPLANATION: explanation\\
\#\# LABEL: label, i.e., "Factual" or "Not Factual"

\begin{table*}[h]
\centering
\begin{tabular}{|p{0.95\textwidth}|}
\hline
\\
Determine whether the given text is factual or not.\\\\
1. Read the input text.\\
2. Evaluate the factual accuracy of the input text based on your training data and knowledge.\\
3. If the input text is factually-accurate, i.e. supported by known information, respond with
"Factual"\\
4. Respond with "Not Factual" if the input text contains even a single inaccuracy.\\
5. Just reply with "Factual" or "Not Factual", do not generate any additional text to the answer.\\\\

\hline
\end{tabular}
\caption{Zero-Shot Prompt for factuality evaluation of a text.}
\label{tab:zs_prompt}
\end{table*}

\begin{table*}[h]
\centering
\begin{tabular}{|p{0.95\textwidth}|}
\hline
\\
Determine whether the given text is factual or not using the provided evidence. If the information is not
present in the evidence, rely on prior knowledge.\\
1. Read the input text.\\
2. Read the evidence if provided.\\
3. Assess whether the input text is factual based on the evidence if present.\\
4. If the evidence are not provided or is insufficient, use your prior knowledge to determine the
factuality.\\
5. Respond with "Not Factual" if the input text contains even a single inaccuracy.\\
6. If the evidence is not related to the text to verify, rely on your prior knowledge to provide the answer.\\
7. Just reply with "Factual" or "Not Factual", do not generate any additional text to the answer.\\\\
\hline
\end{tabular}
\caption{Prompt for factuality evaluation in RAG setting.}
\label{tab:rag_prompt}

\end{table*}

\section{Further details on Evidence Retriever module}
\label{app:retriever}
In this section, we present further details about our evidence retrieval model. To assess the contribution of different components, we performed an ablation study on the retrieval module. All models were trained using the same hyperparameters described in Section \ref{sec:approach_evidence_retrieval}. Results are computed on the corpus $\mathcal{D}$, which contains 2.5 million passages, and evaluated on the validation split of the dataset. After training, our best model achieved a recall at $k=30$ (R@30) of 0.95.

We employed the $E5_{base}$ model \cite{wang2022text}, built upon the $\text{bert-base-uncased}$ \cite{devlin-etal-2019-bert} architecture, with weights initialized from Sentence-Transformers \cite{reimers-gurevych-2019-sentence}. As part of our ablation study, we also trained the $\textit{bert-base-uncased}$ model with the same hyperparameters, achieving a recall of 0.85. This significant performance drop compared to the fully fine-tuned $E5$ demonstrates the effectiveness of the additional pretraining done in $E5$.

Additionally, we experimented with other architectures from the $E5$ family. The $E5_{small}$ model obtained a recall of 0.75, whereas the $E5_{large}$ model slightly outperformed $E5_{base}$, achieving a recall of 0.96. Despite the marginal 1\% performance gain, we opted to use the $E5_{base}$ model in our final system due to the substantial increase in computational resources and training time required by the $E5_{large}$ model, which did not justify the small performance improvement.

The results of all models tested during the ablation study are summarized in Table \ref{tab:retriever-results}, confirming the robustness and efficiency of the $E5_{base}$ model for claim retrieval, balancing performance with computational cost.

\begin{table}[h]
\centering
\begin{tabular}{|l|c|}
\hline
\textbf{Model}       & \textbf{Recall@30} \\ \hline
$E5_{base}$ (without fine-tuning) & 0.52              \\ 
$E5_{base}$           & 0.95              \\ 
$\text{bert-base-uncased}$            & 0.85              \\ 
$E5_{small}$           & 0.75              \\ 
$E5_{large}$           & 0.96              \\ 
\hline
\end{tabular}
\caption{Performance of Different Models on Claim Retrieval Task}
\label{tab:retriever-results}
\end{table}

\section{Details about the employed LLMs}
\label{app:llm_used}
In this section, we detail the models we used in this work.
For the generation of our dataset, we used GPT-4 API, with an approximate cost of \$2000.
As for the open-source models we utilized for the LLM baselines, we used the instruction tuned versions of Mistral \footnote{\url{https://huggingface.co/mistralai/Mistral-7B-Instruct-v0.3}} and LLama 3 \footnote{\url{https://huggingface.co/meta-Llama/Meta-Llama-3-8B-Instruct}}
publicly available on Hugging Face.
For the benchmark evaluation, we utilized the OpenAI API. Specifically, for GPT-4o, we employed the model \textit{GPT-4o-2024-05-13}. 
For the claim-extractor, we use the pre-trained T5-base \footnote{\url{https://huggingface.co/google-t5/t5-base}} as our base model.

\section{Annotation Guidelines}
\label{app:ann_guidelines}

In this section, we illustrate the annotation guidelines employed. Annotators are asked to perform four different tasks related to factuality evaluation. For each task, annotators receive specific guidelines which we report in what follows. As a standard guideline for all tasks, annotators are required to discard instances entirely or partially written in a language other than English. Furthermore, in case of pronominal ambiguity occurring in a given claim, if the human annotator cannot determine, with a high degree of confidence, the noun to which a given pronoun refers, such claim is discarded. Annotators are required to participate in joint sessions to resolve challenges and collaboratively develop agreed-upon solutions.

\subsection{Task 1: Claim Extraction}

\paragraph{Task description}

In this step, you will verify if claims extracted from a given text are accurately represented within the original text. You will receive a 5-sentence passage extracted from Wikipedia, along with corresponding claims pre-extracted by a language model.
Note: a claim denotes an atomic fact, that is, an elementary information unit found in a text, that does not require further subdivision, and that can be checked for its truthfulness.

\paragraph{Annotation Format}

You will be provided with a TSV (Tab-Separated Values) file containing three columns:
\begin{itemize}
    \item Column 1: Identifier (either "text" or "claim {id}")
    \item Column 2: Text or Claim
    \item Column 3: Empty. You have to fill in this column.
\end{itemize}

\paragraph{Annotation Procedure}
\begin{enumerate}
    \item Read the original text and claims thoroughly.
    \item For each claim, determine if it is accurately represented in the original text.
    \item  Place a "v" in the third column if the claim is present in the original text, otherwise mark it with an "x"
\end{enumerate}

\paragraph{Annotation Example}

We report an example of annotated instance in Table \ref{annotated_instance_task1}.

\begin{table*}[]
    \begin{tabularx}{\textwidth}{| m{0.15\textwidth} | X | >{\centering\arraybackslash}m{0.15\textwidth} |}
        \hline
        \textbf{Identifier} & \textbf{Text} & \textbf{Annotation} \\ \hline
        original\_text & This type of meringue is safe to use without cooking. It will not deflate for a long while and can be either used for decoration on pie, or spread on a sheet or baked Alaska base and baked. Swiss meringue is whisked over a bain-marie to warm the egg whites, and then whisked steadily until it cools. This forms a dense, glossy marshmallow-like meringue. It is usually then baked. &  \\ \hline
        claim 1 & Swiss meringue is safe to use without cooking. & v \\ \hline
        claim 2 & Swiss meringue will not deflate for a long while. & v \\ \hline
        claim 3 & Swiss meringue can be used for pie decoration or on a baked Alaska base. & v \\ \hline
        claim 4 & Swiss meringue is whisked over a bain-marie to warm the egg whites. & v \\ \hline
        claim 5 & Swiss meringue is then whisked steadily until it cools. & v \\ \hline
        claim 6 & Swiss meringue forms a dense, glossy, marshmallow-like texture. & v \\ \hline
        claim 7 & Swiss meringue is usually baked after preparation. & v \\ \hline
        claim 8 & Swiss meringue can be mixed with vanilla or chocolate to add flavor. & x \\ \hline
    \end{tabularx}
    \caption{Example of annotated instance in task 1 (claim extraction).}
    \label{annotated_instance_task1}
\end{table*}

\paragraph{Additional Guidelines}

Annotators are required to discard an entire instance, composed of the original text and the corresponding claims, if the original text is not grammatically correct, e.g., if it is syntactically ill-formed, or if it is semantically unclear, that is, if it is formulated in a way that the annotator cannot determine the meaning conveyed either by the entire text or one of its segments. Furthermore, annotators are required to discard sentences which cannot be considered as claims for the purposes of our work, e.g., sentences composed of a single word.

\subsection{Task 2: Claim Falsification}

\paragraph{Task Description}

In this step, you will identify whether a given claim has been altered to introduce unfactual
information.

\paragraph{Annotation Format}

You will receive a pair of claims, where the second claim is an unfactual version of the first

\begin{itemize}
    \item Column 1: The original claim
    \item Column 2: The unfactual claim.
    \item Column 3: Empty. You have to fill in this column.
\end{itemize}

\paragraph{Annotation Procedure}

\begin{enumerate}
    \item Compare the two claims provided.
    \item Determine if the unfactual claim introduces new, untrue information compared to the
original claim.
    \item Mark column 3 with "v" if unfactual information is introduced, otherwise mark it with "x".
\end{enumerate}

\paragraph{Annotation Example}

We report an example of annotated instance in Table \ref{annotated_instance_task3}.

\paragraph{Additional Guidelines}

If the original claim contains a word that is replaced with its hyponym in the candidate nonfactual claim, while the overall meaning of both claims remains unchanged also based on the annotator's world knowledge, then both claims are considered to be semantically equivalent.

\begin{table*}[ht]
    \begin{tabularx}{\textwidth}{| m{0.25\textwidth} | X | >{\centering\arraybackslash}m{0.15\textwidth} |}
        \hline
        \textbf{Identifier} & \textbf{Claim} & \textbf{Annotation} \\ \hline
        claim 1 & The remix in Thank You track was Lassie Come Home. & v \\ \hline
        claim 2 & The remix in Thank You track was not Lassie Come Home. & x \\ \hline
        claim 3 & The Plateau served as a model for colonial capitals. & v \\ \hline
        claim 4 & The Plateau served as a model for other districts. & x \\ \hline
        claim 5 & Christoph Waltz replaced Billy Bob Thornton. & v \\ \hline
        claim 6 & Christoph Waltz replaced Brad Pitt. & x \\ \hline
    \end{tabularx}
    \caption{Example of annotated instance in task 3 (claim falsification).}
    \label{annotated_instance_task3}
\end{table*}

\subsection{Task 3: Factual Text Generation}

\paragraph{Task Description}

In this step, you will assess whether the semantics of claims is preserved in a paraphrased version of the text.

\paragraph{Annotation Format}

You will receive a TSV file with four columns:
\begin{itemize}
    \item Column 1: Identifier (either "paraphrase" or "claim {id}")
    \item Column 2: Text or Claim
    \item Column 3: Empty. You have to fill in this column.
    \item Column 4: Empty. You have to fill in this column.
\end{itemize}

\paragraph{Annotation Procedure}

\begin{enumerate}
    \item Compare each claim with its representation in the paraphrased text.
    \item Determine if its semantics is preserved.
    \begin{itemize}
        \item If it is preserved (regardless of whether it is reported identically in the
paraphrase), place a "v" in the third column.
        \item Use "x" otherwise.
    \end{itemize}
    \item Determine if it is paraphrased.
        \begin{itemize}
            \item If a claim is paraphrased, mark the fourth column with "v".
            \item If not paraphrased (e.g. identical), mark column 4 with "x".
        \end{itemize}
\end{enumerate}

In other words:
\begin{itemize}
    \item <"v", "v"> in the last two columns means that the semantics is preserved and the text is paraphrased (at least one word changed).
    \item <"x", "v"> in the last two columns means that the semantics is NOT preserved but the text is paraphrased.
    \item <"v", "x"> in the last two columns means that the semantics is preserved but the text is NOT paraphrased.
    \item <"v", "x"> in the last two columns means that the semantics is preserved but the text is NOT paraphrased.
    \item <"x", "x"> in the last two columns means that neither the semantics is preserved nor the text is paraphrased (e.g. the claim is omitted). 
\end{itemize}

\paragraph{Annotation Example} 

We report an example of annotated instance in Table \ref{annotated_instance_task3}.

\begin{table*}[]
    \begin{tabularx}{\textwidth}{| l | X | c | c |}
        \hline
        \textbf{Identifier} & \textbf{Text} & \textbf{Semantics Preserved} & \textbf{Paraphrased} \\ \hline
        claim 1 & 'Call Me by Your Name' leads Dorian Award nominations & v & v \\ \hline
        claim 2 & Gregg Kilday authored the article on 10 January 2018 & v & v \\ \hline
        claim 3 & The Hollywood Reporter published the article & v & v \\ \hline
        claim 4 & Article was retrieved on 11 January 2018 & x & x \\ \hline
        claim 5 & The Jameson Empire Awards occurred in 2014 & v & v \\ \hline
        paraphrase & 'Call Me by Your Name' took the lead in Dorian Award nominations. The article, penned by Gregg Kilday, was published by The Hollywood Reporter on January 10, 2018, and accessed the following day. Meanwhile, The Jameson Empire Awards were held back in 2014. & & \\ \hline
    \end{tabularx}
    \caption{Example of annotated instance in task 3 (factual text generation).}
    \label{annotated_instance_task3}
\end{table*}

\paragraph{Additional Guidelines}

If a nearly identical date appears in the factual text and in one claim, annotators should proceed as follows. If the date in the factual text includes the month and year, while the claim specifies the day, month, and year, even if the month and year in the claim coincide with those in the factual text, the semantics conveyed by the claim is considered to be different from that of the factual text.

\subsection{Task 4: Unfactual Text Generation}

\paragraph{Task description}

In this step, you will assess whether all claims, including the unfactual one, are accurately
reflected in a generated unfactual text.

\paragraph{Annotation Format}

You will receive all claims paired with the generated unfactual text.

\begin{itemize}
    \item Column 1: Identifier (either "claim {id}", or "unfactual\_text")
    \item Column 2: Text or Claim
    \item Column 3: Empty. You have to fill in this column.
\end{itemize}

\paragraph{Annotation Procedure}

\begin{itemize}
    \item Review the generated unfactual text along with all claims provided
    \item Determine if all claims are correctly reported in the text (i.e. the factual claims should remain factual and the unfactual claims should be unfactual). Ensure that the text in the “unfactual\_text” field is not modified by the language model to be compliant with the unfactual claim. Paraphrasing in claims is allowed, you should focus on semantics.
    \item Mark column 3 with "v" if the unfactual text corresponds to the claims accurately, otherwise mark it with "x".
\end{itemize}

\paragraph{Annotation Example}

We report an example of annotated instance in Table \ref{annotated_instance_task4}.

\begin{table*}[h]
    \centering
    \begin{tabular}{| m{0.15\textwidth} | m{0.6\textwidth} | m{0.15\textwidth} |}
        \hline
        \textbf{Identifier} & \textbf{Text} & \textbf{Annotation} \\ \hline
        original\_text & In response to crisis, Ottoman statesmen adopted a compliant policy. Abdulmejid's inability to handle the situation heightened discontent regarding the Edict of Tanzimat. \textbf{To enhance European influence, opponents schemed to dethrone Abdulmejid for Abdulaziz.} The planned Kuleli Foundation revolt was thwarted before it could begin on 14 September 1859. Meanwhile, the financial crisis deepened as burdensome foreign debts strained the treasury. &  \\ \hline
        claim 1 & Ottoman statesmen panicked and adopted a policy fulfilling every wish. & v \\ \hline
        claim 2 & Abdulmejid failed to prevent the situation, increasing dissatisfaction with the Edict of Tanzimat. & v \\ \hline
        claim 3 & Opponents planned to replace Abdulmejid with Abdulaziz to enhance European dominance. & v \\ \hline
        claim 4 & The Kuleli Foundation revolt was suppressed before starting on 14 September 1859. & v \\ \hline
        claim 5 & The financial situation worsened, and foreign debts burdened the treasury. & v \\ \hline
        claim 6 & To enhance European influence, opponents schemed to dethrone Abdulmejid for Abdulaziz. & x \\ \hline
    \end{tabular}
    \caption{Example of annotated instance in task 3 (unfactual text generation).}
    \label{annotated_instance_task4}
\end{table*}

\end{document}